\pdfoutput=1

\documentclass[11pt]{article}

\usepackage[final]{acl}

\usepackage{times}
\usepackage{latexsym}

\usepackage[T1]{fontenc}

\usepackage[utf8]{inputenc}

\usepackage{microtype}

\usepackage{inconsolata}

\usepackage{graphicx}

\usepackage{hyperref}
\usepackage{url}
\usepackage{graphicx}
\usepackage{wrapfig}
\usepackage{subcaption}
\usepackage{array}
\usepackage{algorithm}
\usepackage{algorithmic}
\usepackage{multicol}
\usepackage{makecell}
\usepackage{array}
\usepackage{float}
\usepackage{amsmath}

\title{ChuLo: Chunk-Level Key Information Representation \\for Long Document Understanding}

\author{Yan Li \\
  School of Computer Science \\
  The University of Sydney \\
  Sydney, Australia \\
  \texttt{yali3816@uni.sydney.edu.au} \\
  \And
  Soyeon Caren Han \\
  School of Computing and Information Systems \\
  The University of Melbourne \\
  Melbourne, Australia\\
  \texttt{caren.han@unimelb.edu.au} \\
  \AND
  Yue Dai \\
  School of Physics, Mathematics and Computing\\
  The University of Western Australia \\
  Perth, Western Australia, Australia \\
  \texttt{yue.dai@research.uwa.edu.au} \\
  \And
  Feiqi Cao \\
  School of Computer Science \\
  The University of Sydney \\
  Sydney, Australia\\
  \texttt{fcao0492@uni.sydney.edu.au} \\
}

\author{Yan Li$^1$, Soyeon Caren Han$^2$\thanks{Corresponding Author}, Yue Dai$^3$, Feiqi Cao$^1$, \\
$^1$The University of Sydney, $^2$The University of Melbourne,\\ $^3$The University of Western Australia \\
$^1$\texttt{\{yali3816, fcao0492\}@uni.sydney.edu.au}, $^2$\texttt{caren.han@unimelb.edu.au}, \\ $^3$\texttt{yue.dai@research.uwa.edu.au}\\
}

\begin{document}
\maketitle

\begin{abstract}
Transformer-based models have achieved remarkable success in various Natural Language Processing (NLP) tasks, yet their ability to handle long documents is constrained by computational limitations. Traditional approaches, such as truncating inputs, sparse self-attention, and chunking, attempt to mitigate these issues, but they often lead to information loss and hinder the model's ability to capture long-range dependencies. In this paper, we introduce ChuLo, a novel chunk representation method for long document understanding that addresses these limitations. Our ChuLo groups input tokens using unsupervised keyphrase extraction, emphasizing semantically important keyphrase based chunks to retain core document content while reducing input length. This approach minimizes information loss and improves the efficiency of Transformer-based models. Preserving all tokens in long document understanding, especially token classification tasks, is important to ensure that fine-grained annotations, which depend on the entire sequence context, are not lost. We evaluate our method on multiple long document classification tasks and long document token classification tasks, demonstrating its effectiveness through comprehensive qualitative and quantitative analysis. Our implementation is open-sourced on GitHub\footnote{\url{https://github.com/adlnlp/Chulo}}.
\end{abstract}

\section{Introduction}
Transformer-based models \citep{vaswani2017attention}, including LLMs \citep{radford2018improving, radford2019language, brown2020language, ouyang2022training, touvron2023llama, touvron2023llama2, chowdhery2023palm, anil2023palm, dubey2024llama}, have achieved remarkable success across a wide range of Natural Language Processing (NLP) tasks, including Machine Translation, Text Summarization, Text Generation, and Text Classification. A key factor behind their success is the self-attention mechanism, which allows the model to capture long-range dependencies by computing the similarity between any two tokens and aggregating information accordingly. However, this mechanism incurs a quadratic computational cost in terms of both time and space, relative to input length. This computational burden makes it difficult for Transformer-based models to scale to long documents, limiting their application to real-world data with unrestricted document lengths.
To address this challenge, several approaches have been proposed for applying Transformer-based models to long documents while managing computational resources. One of them is truncating, where the model discards content exceeding a predefined input length. For instance, BERT \citep{kenton2019bert} processes up to 512 tokens, and LLaMa \citep{touvron2023llama} handles up to 2048 tokens, with any additional content being ignored. Another one is sparse self-attention, which reduces computational complexity by restricting each query token to attend only to a subset of key tokens \citep{child2019generating,beltagy2020longformer,zaheer2020big,wei2021finetuned,li2023towards}. Lastly, chunking divides long documents into smaller, manageable segments that are processed independently by the model \citep{zhao2021ror,zhang2022summn}. 

While these methods enable Transformer-based models to process long documents, they have limitations. Truncation risks discarding important information that falls beyond the maximum input length. Although more efficient, Sparse attention reduces each token's receptive field, leading to potential information loss from the neglected tokens. Similarly, chunking breaks the input into isolated segments, which can disrupt long-range dependencies critical for a comprehensive understanding of the document.
Preserving all tokens is particularly important in tasks that require fine-grained token-level understanding, such as token classification. In such tasks, dropping tokens can severely impact the accuracy of fine-grained annotations, which often depend on the full context of the document. Therefore, there is a need for methods that can handle long documents efficiently while retaining all key information from the input.

In this paper, we introduce ChuLo, a novel chunk-level key information representation method that addresses these challenges in long document classification and token classification. Our method reduces input length while minimizing information loss by strategically grouping tokens using unsupervised keyphrase extraction. By identifying and emphasizing semantically important tokens, ChuLo ensures that each chunk retains the core content of the document. The resulting chunk representation is used for training Transformer models, with more weight assigned to keyphrases to make them more salient in each chunk. We evaluate ChuLo on various long document classification tasks and long document token classification tasks, demonstrating its effectiveness through competitive results and thorough analysis.

The key contributions of this paper are as follows: 
\textbf{1) Novel Chunk Representation Method}: We introduce ChuLo, a chunk representation method for long document understanding that leverages unsupervised keyphrase extraction to prioritize semantically important information, effectively reducing input length while preserving core content.
\textbf{2) Enhanced Document and Token Classification}: Our proposed method is designed to handle both document-level and token-level tasks, addressing the limitations of existing models in retaining fine-grained annotations and global context in long documents.
\textbf{3) Scalable and Efficient Solution}: ChuLo offers a scalable and efficient approach for long document understanding, making it suitable for various NLP tasks where handling long-range dependencies and context preservation is critical.

\section{Related Work}

\subsection{Long Document Understanding}

Document understanding involves global understanding (e.g., classification) and token-level tasks (e.g., named entity recognition). Transformer-based models face performance issues with long inputs, addressed through input processing and architecture optimization. Input processing methods include truncating tokens beyond the input limit \citep{park2022efficient} and chunking, as seen in Hierarchical Transformer \citep{pappagari2019hierarchical} and RoR \citep{zhao2021ror}, though these often neglect full document context. Architecture optimizations improve efficiency using sparse attention \citep{beltagy2020longformer, zaheer2020big, roy2021efficient} or approximations \citep{peng2021random, wang2020linformer, choromanski2020masked}. Other approaches incorporate RNN concepts, such as cache memory \citep{dai2019transformer, hutchins2022block, li-etal-2023-recurrent}. These methods balance performance and efficiency, highlighting the need to reduce input length effectively.

For document NER, text length is less studied, with recent work addressing low-resource languages \citep{ccetindaug2023named, mengliev2024developing}, complex domains \citep{park2023web, bhattacharya2023improving}, prompt-based methods \citep{wang2023gpt, dagdelen2024structured, hu2024improving}, and multimodal data \citep{yu2023grounded, zhang2023reducing, li-etal-2025-midas}. Our work tackles these challenges\footnote{The summary of Long Document Understanding related works can be found in Appendix \ref{app:related_works}} with a novel chunk representation that preserves semantic information while reducing input length, enhancing both classification and token-level tasks.

\subsection{Unsupervised Keyphrase Extraction}
Unsupervised keyphrase extraction identifies representative phrases to summarize content without labelled data \citep{hasan-ng-2014-automatic}. Methods include statistics-based (e.g., TfIdf \citep{el2009kp}, co-occurrence \citep{liu2009clustering}, and context statistics \citep{campos2020yake, won2019automatic}), graph-based (e.g., TextRank \citep{mihalcea2004textrank} and its variants \citep{wan2008single, bougouin2013topicrank, florescu2017positionrank, yu2018wikirank}), and embedding-based approaches (e.g., EmbedRank \citep{bennani2018simple}, SIFRank \citep{sun2020sifrank}, and PromptRank \citep{kong-etal-2023-promptrank}). While effective, these methods prioritize phrase extraction and ranking over improving downstream tasks. Our work integrates keyphrase extraction with chunk representation to enhance long document understanding.

\begin{figure*}[th]
\vspace{-0.1in}
\includegraphics[width=1.0\linewidth]{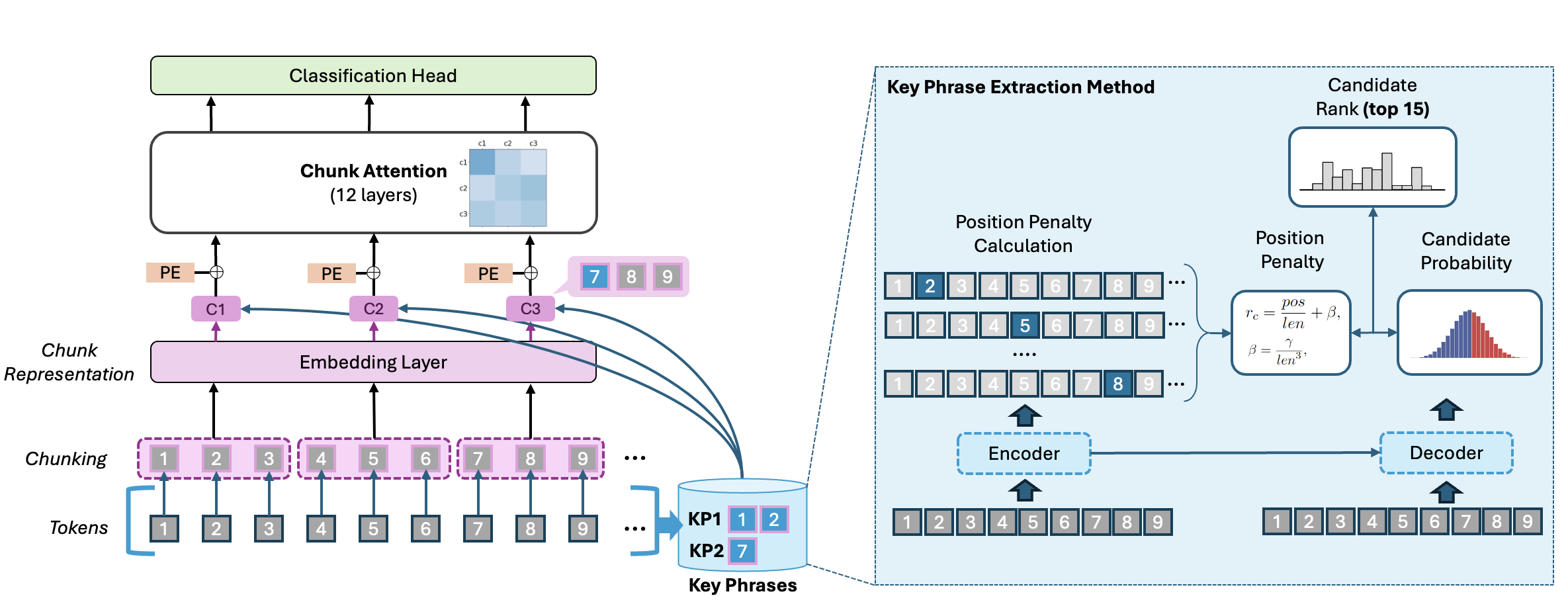}
\vspace{-0.1in}
\caption{The Overall ChuLo Framework proposed in this paper. Each chunk is surrounded by a pink box. $C_1$ ... $C_n$ represents the chunk representation.}
\label{fig:framework}
\vspace{-5mm}
\end{figure*}

\section{ChuLo}
We propose ChuLo, a novel chunk representation method that enhances long document understanding by reducing input length while preserving semantic content. Unlike existing approaches like truncation and standard chunking, ChuLo minimizes information loss and maintains contextual dependencies. The method segments documents into non-overlapping chunks, integrates key semantic information using unsupervised keyphrase extraction, and assigns higher weights to keyphrase tokens in chunk representations. These enriched chunks train a Transformer-based chunk attention module, enabling efficient processing of long documents while retaining global and local context. Further details are provided in subsequent subsections, with the framework illustrated in Figure \ref{fig:framework}.

\subsection{Document Input Chunking}
To effectively manage long document inputs, we employ a chunking strategy that reduces input length while preserving all relevant information.
Common approaches to long document processing, such as truncation and sparse attention, either disregard parts of the document\citep{lewis2020bart,park2022efficient} or restrict the receptive field of individual tokens \citep{beltagy2020longformer,zaheer2020big,brown2020language}, resulting in potential information loss.
Our approach mitigates these issues by segmenting the document into non-overlapping chunks before feeding them into the model. It enables complete self-attention among chunks, ensuring that all information is retained and enabling the model to process larger portions of the document context. Specifically, we first tokenize the document $D = (t_0, t_1, \dots, t_{l_{D}-1})$ and divide it into fixed-length chunks $C_D = (C_0, C_1, \dots, C_{m-1})$, where $l_{D}$ is the number of the tokens, each chunk $C$ consists of $n$ tokens, and $m= \lceil {\frac{l_{D}}{n}} \rceil$ is the number of chunks. The incomplete chunks will be padded with the [PAD] tokens. The chunk size $n$ is a hyper-parameter controlling the degree of input length reduction. By grouping tokens this way, we maintain the integrity of the input content while alleviating the computational burden associated with processing long sequences.

\subsection{Semantic Key Information Extraction}
The fundamental reason for extracting keyphrases from the chunks, as defined in the document chunking step, is to maintain the integrity of the document's semantic content while reducing input length. During chunking, the document is divided into smaller segments, which can inadvertently distribute important semantic information unevenly across chunks or even cause it to be diluted. Simply treating each chunk equally may lead to overlooking critical context essential for accurate document classification and token-level understanding.
Identifying and highlighting critical phrases within these chunks ensures that the most relevant information is preserved and emphasized, allowing the model to focus on the core content even within a limited input space. This compensates for the information fragmentation caused by chunking and guides the Transformer’s attention mechanism to prioritize the most informative parts of the text, enhancing the model’s ability to capture the document’s overall meaning and relationships. Thus, extracting keyphrases from chunks is crucial for bridging the gap between document segmentation and semantic coherence, ultimately improving the effectiveness of the chunk-based representation for long document understanding.
To achieve this, we extract semantically important keyphrases to identify the core content of the entire document. Since document understanding, such as document classification or token classification, inherently involves semantic understanding, it is crucial to highlight the most informative parts of the text to create meaningful chunk representations. By making the extracted keyphrases more salient, we can effectively emphasize the content that contributes most to the document’s overall meaning. Hence, we employ unsupervised keyphrase extraction methods, ensuring our approach remains adaptable across diverse domains without requiring annotated data. Building on the principles of PromptRank \citep{kong-etal-2023-promptrank}, we adapt and enhance its template-based approach to prioritize keyphrases that are contextually significant across the entire document. Our modified strategy, the Semantic Keyphrase Prioritization (SKP) Algorithm, leverages prompts to assess the importance of each candidate keyphrase, ensuring that semantically crucial information is highlighted for downstream document understanding. The details of this process are provided in Appendix \ref{app:algorithm}. In this way, our method emphasizes keyphrases during chunk representation, making critical semantic information more salient. Consequently, our method bridges the gap between document segmentation and semantic coherence by ensuring that key content is preserved and highlighted within the entire document, despite input length constraints.


\subsection{Chunk Representation Production}
After extracting the semantically significant keyphrases, we construct a chunk representation that preserves and highlights this key information, ensuring that the chunk retains the core semantic content of the document. While chunking helps reduce the input length, it may also result in an uneven distribution of meaningful content across chunks. Thus, it is crucial to re-emphasize the importance of these keyphrases within the chunk to maintain semantic integrity. Our approach dynamically adjusts the representation of each chunk by assigning greater importance to keyphrase tokens, enabling the model to focus on the most relevant content during downstream processing.
To achieve this, we label the tokens corresponding to the extracted keyphrases in the original text as keyphrase tokens $T_k$, while other tokens are labelled as non-keyphrase tokens $T_{nk}$. Then, we feed these chunked tokens $t$ into the embedding layer to obtain their embeddings. The chunk embedding $\boldsymbol{c}$ is then computed using a weighted average of these token embeddings, as defined in Formula \ref{equ:chunkemb}:


\begin{equation}
    \begin{cases}
        w_t = \begin{cases}
        a, \text{t is $T_k$}\\b, \text{t is $T_{nk}$}
        \end{cases} \\ 
       \boldsymbol{c} = \frac{\sum{w_t*\boldsymbol{t}}}{\sum{w_t}}
    \end{cases}
    \label{equ:chunkemb}
\end{equation}


Here, $w_t$ represents the weight assigned to each token $t$ in the chunk, where $a$ and $b$ are hyperparameters with $a > b.$
$\boldsymbol{t}$ denotes the embedding of token $t$, and $\boldsymbol{c}$ is the resulting chunk embedding that captures the weighted importance of keyphrase and non-keyphrase tokens. By assigning a higher weight $a$ to keyphrase tokens, we ensure that the resulting chunk representation emphasizes the most critical information while maintaining a compact input length.
Finally, the chunk embeddings are fed into the Transformer-based model, allowing it to effectively leverage the enhanced chunk representations during long document classification or token-level classification tasks. This method not only preserves the semantic coherence of the document but also allows the model to retain meaningful context and relationships, ultimately enhancing its performance on long document tasks.

\subsection{Chunk Representation Training}
In this final step, we train a Transformer-based model using our keyphrase-enhanced chunk representations to effectively incorporate the core semantic content of the document. We selected BERT-base according to Table \ref{tab:backbone ablation}. By emphasizing key information in the chunk embeddings, we ensure that the model can focus on the most relevant aspects of the text, thereby improving its ability to handle long document inputs without losing critical context.
We leverage a Transformer-based backbone model, which is used to initialize the weights of the chunk attention module, as illustrated in Figure~\ref{fig:framework}. This chunk attention module is designed to capture the intricate contextual relationships among chunks while retaining the influence of keyphrases. By doing so, the module can better understand local and global semantic patterns, enabling the model to perform robustly across various long document tasks. The chunk embeddings are processed through the chunk attention module to produce refined contextual representations, which are then fed into a classification head to generate the final predictions.
Our framework, ChuLo, is adaptable to any transformer-based architecture, from pretrained to large language models, making it versatile for tasks involving long document understanding. Through integrating keyphrase-enhanced chunk representations, the model achieves superior performance in both document classification and token-level tasks, highlighting the effectiveness of our approach in leveraging semantic information to tackle the challenges associated with long document processing.

\section{Experiments Set-up}
We evaluate ChuLo on document and token classification tasks to highlight our motivation. While document classification primarily requires global contextual understanding, token classification tasks test the model's ability to retain and utilize detailed token-level information within long documents. We compare it with BERT \citep{kenton2019bert} and BERT variants \citep{park2022efficient}, Longformer \citep{beltagy2020longformer}, ToBERT \citep{pappagari2019hierarchical}, CogLTX \citep{ding2020cogltx}, ChunkBERT \citep{jaiswal2023breaking}, and instructions with LLMs, GPT4o\footnote{\url{https://openai.com/index/hello-gpt-4o/}} and Gemini1.5pro\footnote{\url{https://deepmind.google/technologies/gemini/pro/}}. \textbf{Baselines Details} are listed in Appendix \ref{app:baseline models}.

\textbf{Datasets: } We conduct experiments using three(3) datasets for long document classification and two(2) for long document token classification. For document classification, we use HP\citep{kiesel2019semeval}, LUN \citep{rashkin2017truth}, and Eurlex57k \citep{chalkidis2019large}. These datasets vary in average document length from 707 to 1147 tokens, enabling us to assess performance on documents of different lengths and complexities. Further details on dataset statistics and splits are available in Appendix \ref{sec:Appendix:Dataset statistics}. 
\textbf{1) HP} (Hyperpartisan News Detection) evaluates the classification of news articles based on hyperpartisan argumentation \citep{kiesel2019semeval}. We use the ‘byarticle’ version, which contains 238 hyperpartisan and 407 non-hyperpartisan articles. The same train-test split as in \citep{beltagy2020longformer} is adopted.
\textbf{2) LUN} uses for unreliable news source classification, this dataset includes 17,250 articles from satire, propaganda, and hoaxes \citep{rashkin2017truth}. Our goal is to predict the source type for each article.
\textbf{3) Eurlex57k} consists of 47,000 English EU legislative documents with 4,271 EUROVOC concepts. We also include a simulated \textbf{Inverted-Eurlex57k} version, where the header and recitals are moved to the end, compelling the model to read the entire document for key information.
For token classification, we use GUM and CoNLL-2012 for Named Entity Recognition (NER) tasks:
\textbf{1) GUM} (Georgetown University Multilayer) is a richly annotated collection of 235 documents across various genres such as academic texts, news, fiction, and interviews \citep{lin-zeldes-2021-wikigum}. GUM’s various linguistic styles and structures make it an excellent benchmark for assessing token-level understanding in lengthy documents, ensuring that the model captures complex entity relationships over extended contexts.
\textbf{2) CoNLL}-2012 originally designed for coreference resolution, and is adapted for NER in our work \citep{pradhan-etal-2013-towards}. We convert the data to a document-level format and select the top-k longest documents in each split, emphasizing the model’s ability to process extended text sequences for token classification tasks.

\textbf{Metrics and Implementation: } For the HP and LUN datasets, we use \textbf{accuracy} as the evaluation metric, while for Eurlex57k, Inverted Eurlex57k, GUM, and CoNLL-2012, we adopt \textbf{micro F1}. These metrics are chosen to maintain consistency with prior work and facilitate direct comparison.
Regarding implementation, we provide key details here, with the complete setup in Appendix \ref{app:imp. details}. We use CrossEntropy loss for training on the Hyperpartisan, LUN, CoNLL and GUM datasets, and Binary CrossEntropy loss for the Eurlex57k and Inverted Eurlex57k datasets. All models are optimized using the AdamW optimizer, and training employs early stopping based on the respective validation metric, with a patience threshold set to 10 epochs. A learning rate search is conducted for each experiment to ensure optimal model performance for comparison. Top-n value is set to 15\footnote{We tested with different n values, but 15 was generally better in most datasets}.

\section{Results}
\subsection{Document Classification}
We evaluate the effectiveness of our ChuLo by comparing it with fine-tuned PLMs and previously published baselines \citep{park2022efficient, jaiswal2023breaking} on several benchmark datasets: HP, LUN, EURLEX57K, and Inverted EURLEX57K. The comparative results are summarized in Table \ref{tab:overal_performance_comparison}, with input configurations provided in Table \ref{tab:usage_of_input} and detailed descriptions available in Appendix \ref{app: baseline input}. 
Our method demonstrates clear superiority on three of the four datasets, achieving a significant improvement of 6.43\% accuracy on the LUN dataset compared to the second-best model, BERT. This marked improvement presents ChuLo’s ability to capture comprehensive document context through its keyphrase-based chunk representation, despite using only 512 input embeddings. The results suggest that our method effectively mitigates the limitations of traditional truncation and chunking strategies by preserving critical semantic information, which contributes to higher classification accuracy. On the EURLEX57K and Inverted EURLEX57K datasets, ChuLo achieves consistent performance gains over baselines, further validating its capability to handle long documents efficiently. In these datasets, which have hierarchical labels and require understanding complex semantic structures, our model benefits from enhanced chunk representations that emphasize key content. This allows ChuLo to capture document semantics better, even when compared to models that can process larger input lengths.
While our model delivers competitive results on the HP dataset, it trails behind Longformer by a slight margin of 0.0031 in accuracy. This marginal difference corresponds to only one additional correctly classified instance out of a total of 65 test samples. 

\setlength{\tabcolsep}{9pt}
\begin{table}[t]
\scriptsize
\centering
\begin{tabular}{l|cccc}
\noalign{\hrule height 0.8pt}
\textbf{Model} & \textbf{HP} & \textbf{LUN} & \textbf{EUR} & \textbf{I-EUR} \\
\hline
BERT & 0.9200 & \underline{0.5797} & 0.7309 & 0.7053 \\
ToBERT & 0.8954 & 0.3697 & 0.6757 & 0.6731 \\
CogLTX & 0.9477 & - & 0.7013 & 0.7080 \\
Longformer & \textbf{0.9569} & 0.5552 & 0.5453 & 0.5647 \\
BERT+TextRank$^{\dagger}$ & 0.9115 & 0.4880 & 0.7287 & 0.7130 \\
BERT+Random$^{\dagger}$  & 0.8923 & 0.3015 & \underline{0.7322} & \underline{0.7147} \\
ChunkBERT & 0.9300 & - & 0.6494 & 0.6294 \\
\hline
\textbf{Ours} & \underline{0.9538} & \textbf{0.6440} & \textbf{0.7332} & \textbf{0.7244} \\
\noalign{\hrule height 0.8pt}
\end{tabular}
\caption{Document classification Result. Following previous work, we use accuracy for HP and LUN, and micro F1 for other datasets. Results for LUN are obtained by our own experiment based on provided baseline codes and methods, while baseline results for the other datasets are from previous work\citep{park2022efficient, jaiswal2023breaking}. $^{\dagger}$ are the BERT variants proposed by \cite{park2022efficient}. The best performance for each dataset is bolded while the second best is underscored, and we can see that our final model, a BERT-based backbone, generally outperforms other baselines across all datasets by achieving the best or second-best.}
\label{tab:overal_performance_comparison}
\vspace{-10pt}
\end{table}


\begin{table}[ht]
\vspace{-2pt}
\setlength\tabcolsep{8pt}
\scriptsize
\centering
\begin{tabular}{l|c}
\noalign{\hrule height 0.8pt}
\textbf{Model} & \textbf{The Usage of Input} \\
\hline
BERT & F-512 tokens\\
ToBERT & All \\
CogLTX & S-512 tokens\\
Longformer & F-4096 tokens \\
BERT+TextRank & F-512 + S-512 tokens\\
BERT+Random & F-512 + S-512 tokens \\
ChunkBERT & All \\
\hline
\textbf{Ours} & All (512*Chunk Size) \\
\noalign{\hrule height 0.8pt}
\end{tabular}
\caption{The usage of the input content in the experiments.``F-512`` and ``F-4096`` means the first 512 tokens and the first 4096 tokens, ``S-512`` means the selected 512 tokens.}
\label{tab:usage_of_input}
\vspace{-5pt}
\end{table}

Interestingly, for the other datasets, Longformer underperforms compared to models like BERT variants or CogLTX, which use the first 512 tokens and focus on selecting key sentences. This observation indicates that unfiltered additional content can introduce noise, negatively impacting classification accuracy. In contrast, ChuLo expands the input content and strategically emphasizes key semantic elements during chunk representation. This approach mitigates noise interference, ensuring that only the most relevant information is retained and highlighted. 
Overall, the results confirm that ChuLo consistently outperforms standard PLM baselines and existing chunking methods in long document classification tasks. Its ability to retain and emphasize key semantic content, while efficiently handling long inputs, makes it a robust solution for various document classification challenges.

\subsection{Longer Document Classification}
To further validate the robustness of our model, we evaluate its classification performance across various document length ranges, with a particular focus on longer documents. For this analysis, we consider the documents with more than 1024 tokens and more than 2048 tokens in the test set. 
We use Longformer 
and off-the-shelf LLMs, GPT4o and Gemini1.5 pro for comparison. 
As shown in Table \ref{tab:Accuracy length intervals}, our model consistently outperforms others on longer documents in the LUN dataset. Specifically, for documents exceeding 2,048 tokens, ChuLo maintains a higher accuracy compared to all baselines, demonstrating its capacity to handle lengthy inputs effectively. This performance gain can be attributed to our chunk representation’s emphasis on keyphrases, which preserves crucial semantic content even when document length increases.
On the HP dataset, ChuLo and Longformer achieve perfect accuracy (1.0) for documents longer than 2,048 tokens. However, for shorter documents (more than 1,024 tokens), ChuLo surpasses Longformer. This improvement is likely due to our chunk representation strategy, which selectively highlights key content rather than averaging information across the entire document. As a result, ChuLo maintains high semantic fidelity, leading to better overall performance even with condensed text inputs.

\begin{table}[ht]
\centering
\scriptsize
\setlength{\tabcolsep}{6pt}
\begin{subtable}{\linewidth}\centering
    \begin{tabular}{p{0.3\linewidth}|p{0.15\linewidth}p{0.15\linewidth}p{0.15\linewidth}}
    \noalign{\hrule height 0.8pt}
    \textbf{LUN} & \textbf{All(2250)} & \textbf{1024(243)} & \textbf{2048(49)} \\
    \hline
    Longformer & 0.5552 & 0.4062 & 0.5306 \\
    GPT4o & -  & - & 0.7143 \\
    Gemini1.5pro & -  & - & 0.6531 \\
    \hline
    \textbf{Ours}  & \textbf{0.6741} & \textbf{0.5911} & \textbf{0.7959} \\
    \noalign{\hrule height 0.8pt}
    \end{tabular}
    \caption{LUN dataset}
\end{subtable}
\vspace{0.5cm} 
\begin{subtable}{\linewidth}\centering
    \begin{tabular}{p{0.3\linewidth}|p{0.15\linewidth}p{0.15\linewidth}p{0.15\linewidth}}
    \noalign{\hrule height 0.8pt}
    \textbf{HP} & \textbf{All(65)} & \textbf{1024(28)} & \textbf{2048(9)} \\
    \hline
    Longformer & \textbf{0.9538}  & 0.8929 & \textbf{1.000} \\
    GPT4o & -  & - & 0.8889 \\
    Gemini1.5pro & -  & - & 0.7778 \\
    \hline
    \textbf{Ours}  & \textbf{0.9538} & \textbf{0.9286} & \textbf{1.000} \\
    \noalign{\hrule height 0.8pt}
    \end{tabular}
    \caption{HP dataset}
\end{subtable}
\vspace{-25pt}
\caption{Document classification results for comparison on documents of different lengths: all documents in the test set, the subset of documents longer than 1024 tokens, and longer than 2048 tokens respectively. Values in brackets indicate the number of documents for each specific document set. The best performance (Accuracy) for each document set is bolded.}
\label{tab:Accuracy length intervals}
\vspace{-10pt}
\end{table}

We also benchmarked against newly released LLMs, GPT-4o and Gemini 1.5 Pro, using longer document inputs for both the LUN and HP datasets. On LUN, GPT-4o achieved an accuracy of 0.7143 and Gemini 1.5 Pro scored 0.6531, both surpassing Longformer. However, ChuLo achieved the highest accuracy of 0.7959, showcasing its superiority in handling long documents with diverse content. On the HP dataset, GPT-4o (0.8889) and Gemini 1.5 Pro (0.7778) performed worse than Longformer and ChuLo, both of which achieved a perfect accuracy of 1.0 on the longer documents. This highlights ChuLo’s robustness and consistency in classifying documents with varying length, even compared to advanced language models. The prompt and response samples are in Section \ref{sec:prompt} and \ref{app:cases}. Overall, these results demonstrate that ChuLo not only outperforms standard PLM baselines and chunking methods on long documents but also remains competitive against the latest large language models. By prioritizing key semantic elements and managing document length, ChuLo maintains stable performance across varying input lengths.

\begin{table}[htp]
\setlength{\tabcolsep}{8pt}
\centering
\scriptsize
\vspace{-5pt}
\begin{tabular}{p{0.3\linewidth}|>{\centering\arraybackslash}p{0.225\linewidth}>{\centering\arraybackslash}p{0.225\linewidth}}
\noalign{\hrule height 0.8pt}
\textbf{Model}  & \textbf{CoNLL} & \textbf{GUM}\\
\hline
Longformer (4096)    & 0.5560 & 0.9427\\
BigBird (4096)       & 0.5553 & 0.9418\\
GPT4o      & 0.2290 & 0.3231\\
Gemini1.5    & 0.3036 & 0.3262\\
\hline
\textbf{Ours (All)} &  \textbf{0.9334} & \textbf{0.9555}\\
\noalign{\hrule height 0.8pt}
\end{tabular}
\caption{Results on token classification tasks. The best performance for each dataset is bolded, and our model achieves the best on both datasets.}
\label{tab:overal_performance_token_cls_comparison}
\vspace{-10pt}
\end{table}

\subsection{Token Classification}
To further demonstrate the effectiveness of our chunk representation method, we evaluated it on a token-level classification task—specifically, Named Entity Recognition (NER) using long documents. We compared our model against two state-of-the-art long-document pre-trained models, Longformer \citep{beltagy2020longformer} and BigBird \citep{zaheer2020big}, as well as newly released large language models, GPT-4o and Gemini 1.5 Pro. 
As shown in Table \ref{tab:overal_performance_token_cls_comparison}, our model consistently outperforms Longformer, BigBird and LLM models on the NER tasks, particularly on the CoNLL, where document lengths often exceed the input limitations of these baseline models. To leverage the broader context captured by our chunk representation, we integrate a BERT-decoder module that utilizes the enhanced chunk embeddings to predict token labels more accurately. This configuration allows the model to maintain a global understanding of the document while focusing on the local dependencies necessary for precise token classification.
All baselines struggle with these longer inputs due to their limited capacity for handling extensive sequences. In contrast, our method’s ability to encode the entire document’s context through keyphrase-based chunk representations enables it to achieve higher accuracy in recognizing and classifying named entities. This is particularly evident in cases where long-distance dependencies and contextual nuances play a critical role in determining the correct labels.
Overall, the results indicate that our model's chunk representation not only enhances performance on document-level classification tasks but also proves highly effective for token-level tasks such as NER, 
validating its application in downstream tasks that require a detailed and comprehensive understanding of long document tokens.




\begin{table}[!ht]
\scriptsize
\centering
\begin{subtable}{\linewidth}\centering
    \setlength{\tabcolsep}{6.1pt}
    \begin{tabular}{l|ccccc}
    \noalign{\hrule height 0.8pt}
    \textbf{CoNLL} & \textbf{ALL (20)} & \textbf{> 2048 (17)} & \textbf{> 4096(6)} & \textbf{> 8192 (2)}\\
    \hline
    Longformer   & \underline{0.5560} & \underline{0.5268} & \underline{0.3156} & \underline{0.3116} \\
    BigBird   & 0.5553 & 0.5261 & 0.3145 & 0.3106 \\
    GPT4o   & 0.2290 & 0.2217 & 0.1252 & 0.0282 \\
    Gemini 1.5   & 0.3036 & 0.2633 & 0.1652 & 0.0584 \\
    
    \hline
    \textbf{Ours}   & \textbf{0.9334} & \textbf{0.9325} & \textbf{0.9287} & \textbf{0.9206}\\
    \noalign{\hrule height 0.8pt}
    \end{tabular}
    \caption{Results on CoNLL dataset.}
    \label{tab:Conll Microf1 length intervals}
\end{subtable}

\begin{subtable}{\linewidth}\centering
    \setlength{\tabcolsep}{9.2pt}
    \begin{tabular}{l|cccc}
    \noalign{\hrule height 0.8pt}
    \textbf{GUM} & \textbf{ALL (26) - > 512} & \textbf{> 1000(8)} & \textbf{> 1042(6)} \\
    \hline
    Longformer   & \underline{0.9427}  & \underline{0.9427} & \underline{0.9439}\\
    BigBird   & 0.9418  & 0.9417 & 0.9426\\
    GPT4o   & 0.3231  & 0.3018 & 0.2808\\
    Gemini 1.5   & 0.3262  & 0.3093 & 0.3215\\
    \hline
    \textbf{Ours}   & \textbf{0.9555}  & \textbf{0.9558} & \textbf{0.9574}\\
    \noalign{\hrule height 0.8pt}
    \end{tabular}
    \caption{Results on GUM dataset.
    }
    \label{tab:Gum Microf1 length intervals}
    \vspace{-5pt}
\end{subtable}
\caption{NER results for comparison on documents of different lengths. \text{>}\textit{number} represents the documents longer than the number, with the values in brackets indicating the corresponding counts for the documents. The best performance (Micro F1) is bolded and the second best is underscored, and our model consistently outperforms all the baselines for each document set.}
\vspace{-5pt}
\end{table}

\begin{figure*}[!htp]
    \begin{subfigure}[b]{0.47\linewidth}   
        \includegraphics[width=\textwidth]{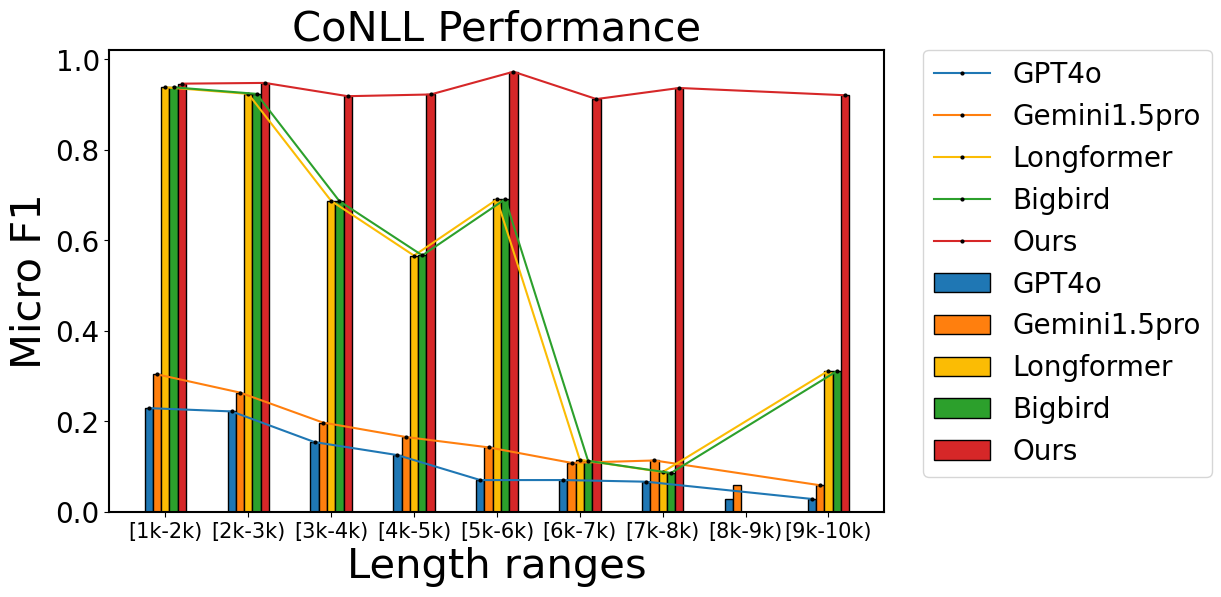} 
        \caption{CoNLL Performance (Range: 1798 to 9778)}
        \label{fig:conll_results}
    \end{subfigure}
    \hfill
    \begin{subfigure}[b]{0.47\linewidth}  
        \includegraphics[width=\textwidth]{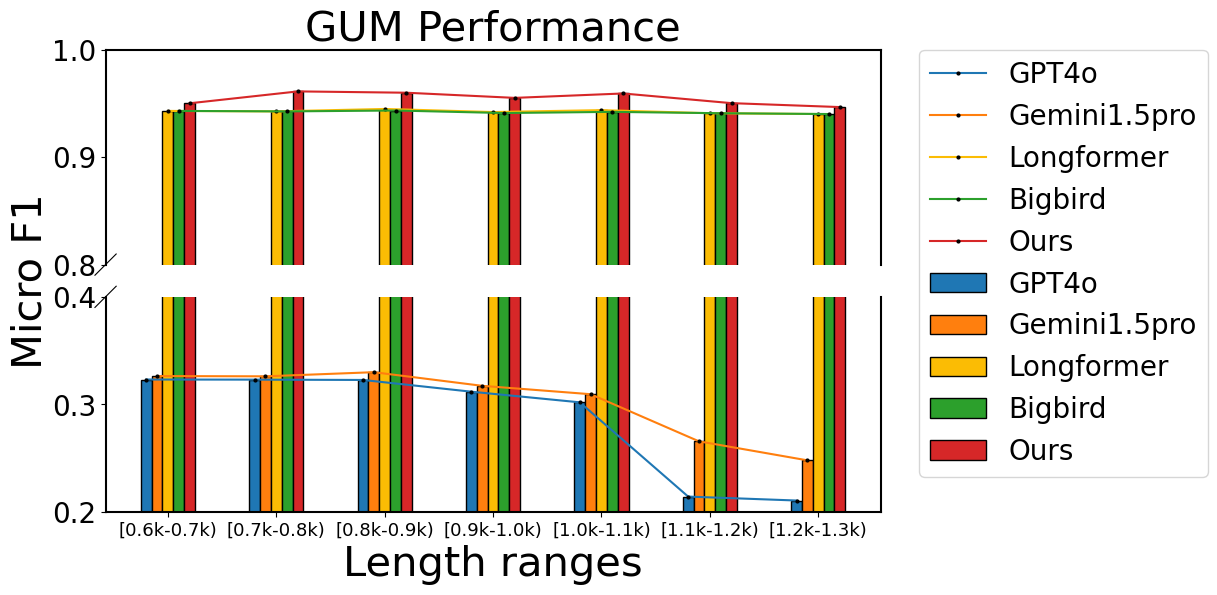} 
        \caption{GUM Performance (Range: 628 to 1281)}
        \label{fig:gum_results}
    \end{subfigure}
    \caption{Comparison of performance in different length ranges for CoNLL and GUM datasets. Values of brackets includes the min and max length of each dataset.}
    \label{fig:combined_results}
    \vspace{-10pt}
\end{figure*}

\subsection{Token Classification in Longer Documents}
We further analyze the NER performance across different document length ranges. As presented in Table \ref{tab:Conll Microf1 length intervals} and Table \ref{tab:Gum Microf1 length intervals}, we report the number of documents exceeding specific length thresholds and their corresponding performance metrics. On the CoNLL, as document lengths exceed the maximum input capacities of Longformer and BigBird, both models exhibit significant performance drops to 31.56\% and 31.45\%, respectively. In contrast, our model experiences a minimal decrease of 1.28\%, showcasing its resilience and effectiveness in handling long sequences. For the GUM, where all document lengths are within the acceptable range for these models, performance remains stable across all models, with our approach consistently achieving the best results.
Figures \ref{fig:conll_results} and \ref{fig:gum_results} visualize the performance breakdown across varying length ranges. For the CoNLL, our model maintains high performance in all length intervals, while Longformer and BigBird exhibit comparable performance within the [1k-2k) range but degrade significantly for longer texts, even for documents that do not exceed their maximum input length. This discrepancy suggests that the uneven distribution of document lengths in their pretraining corpora may lead to inconsistent performance on longer sequences. In contrast, our model’s ability to compress the entire document into 512-length chunks for the decoder enables it to leverage complete contextual information, resulting in better stability and accuracy even on longer documents.
For the GUM, where document lengths are shorter (up to 1.3k tokens), our model consistently outperforms Longformer and BigBird in all intervals. The stable performance of all models on GUM aligns with the results on CoNLL, further confirming that our approach’s chunk representation is particularly effective when documents reach lengths that exceed the standard input capacities of the baselines.
These results underscore the effectiveness of our chunk representation, which emphasizes keyphrase information, for coarse-grained document classification and fine-grained token-level classification tasks like NER. The ability to maintain performance across varying document lengths highlights the importance of incorporating global contextual information in NER tasks—a largely underexplored aspect. Additionally, off-the-shelf LLMs such as GPT-4o and Gemini 1.5 Pro show suboptimal performance on NER tasks without fine-tuning, and their performance deteriorates further as document length increases. This indicates that, despite their advancements, LLMs still require substantial optimization for effective long document understanding.

\subsection{Prompt Method}
\label{sec:prompt}
We employed zero-shot prompting with large language models (LLMs), specifically Gemini 1.5 Pro and GPT4o, in our experiments. The prompts used for each dataset are detailed in Table \ref{tab:app_prompts_doc_c} and \ref{tab:app_prompts_token_c}:

\begin{table}[htp]
\renewcommand{\arraystretch}{1.5}
\setlength{\tabcolsep}{2pt}
\scriptsize
\centering
\begin{tabular}{p{1.2cm}|p{6cm}}
\noalign{\hrule height 0.8pt}
\textbf{Dataset} & \multicolumn{1}{c}{\textbf{Prompt}} \\
\hline
LUN   &  Task Definition: You are provided with a news article. Your task is to classify the article into one of the following categories: "Satire” "Hoax” or "Propaganda” Respond only with the appropriate category. The news is: [\{input\}]. \\
\hline
HP   & Task Definition: You are provided with a news article. Your task is to classify whether the article is hyperpartisan. Respond only with "True” if the news is hyperpartisan or "False” if it is not. The news is: [\{input\}]. \\
\noalign{\hrule height 0.8pt}
\end{tabular}
\caption{The prompt we used for each dataset in our experiments.}
\label{tab:app_prompts_doc_c}
\end{table}

\begin{table*}[htp]
\renewcommand{\arraystretch}{1.5}
\setlength{\tabcolsep}{2pt}
\scriptsize
\centering
\begin{tabular}{p{1.2cm}|p{12cm}}
\noalign{\hrule height 0.8pt}
\textbf{Dataset} & \multicolumn{1}{c}{\textbf{Prompt}} \\
\hline
CoNLL  & In the task of Named Entity Recognition, the B-, I-, and O- prefixes are commonly used to annotate slot types, indicating the boundaries and types of slots. These labels typically represent:
        B- (Begin): Signifies the beginning of a slot, marking the start of a new slot.
        I- (Inside): Represents the interior of a slot, indicating a continuation of the slot.
        O (Outside): Denotes parts of the input that are not part of any slot.
        For instance, in a sentence where we want to label a "date" slot, words containing date information might be tagged as "B-date" (indicating the beginning of a date slot), followed by consecutive words carrying date information tagged as "I-date" (indicating the continuation of the date slot), while words not containing date information would be tagged as "O" (indicating they are outside any slot).

Definition: In this task, you are given a conversation, where the words spoken by a person are shown as a list. Your job is to classify the words in the following conversation into one of the 37 different entities. The entities are: "O", "B-PERSON", "I-PERSON", "B-NORP", "I-NORP", "B-FAC", "I-FAC", "B-ORG", "I-ORG", "B-GPE", "I-GPE", "B-LOC", "I-LOC", "B-PRODUCT", "I-PRODUCT", "B-DATE", "I-DATE", "B-TIME", "I-TIME", "B-PERCENT", "I-PERCENT", "B-MONEY", "I-MONEY", "B-QUANTITY", "I-QUANTITY", "B-ORDINAL", "I-ORDINAL", "B-CARDINAL", "I-CARDINAL", "B-EVENT", "I-EVENT", "B-WORK\_OF\_ART", "I-WORK\_OF\_ART", "B-LAW", "I-LAW", "B-LANGUAGE", "I-LANGUAGE". Only output entities. And the entity types should be output as a list without any explanation. The input is [\{input\}]. \\
\hline
GUM  & In the task of Named Entity Recognition, the B-, I-, and O- prefixes are commonly used to annotate slot types, indicating the boundaries and types of slots. These labels typically represent:
        B- (Begin): Signifies the beginning of a slot, marking the start of a new slot.
        I- (Inside): Represents the interior of a slot, indicating a continuation of the slot.
        O (Outside): Denotes parts of the input that are not part of any slot.
        For instance, in a sentence where we want to label a "date" slot, words containing date information might be tagged as "B-date" (indicating the beginning of a date slot), followed by consecutive words carrying date information tagged as "I-date" (indicating the continuation of the date slot), while words not containing date information would be tagged as "O" (indicating they are outside any slot).

Definition: In this task, you are given a conversation, where the words spoken by a person are shown as a list. Your job is to classify the words in the following conversation into one of the 37 different entities. The entities are: "I-abstract", "B-object", "B-place", "I-substance", "I-time", "I-place", "B-time", "B-abstract", "I-person", "B-plant", "B-substance", "I-animal", "B-organization", "I-event", "B-person", "B-event", "I-plant", "I-organization", "O", "I-object", "B-animal". Only output entities. And the entity types should be output as a list without any explanation. The input is [\{input\}]. \\
\noalign{\hrule height 0.8pt}
\end{tabular}
\caption{The prompt we used for each dataset in our experiments.}
\label{tab:app_prompts_token_c}
\end{table*}

\begin{figure}[!h]
    \centering
    \begin{subfigure}[b]{1.0\linewidth}   
        \includegraphics[width=\textwidth]{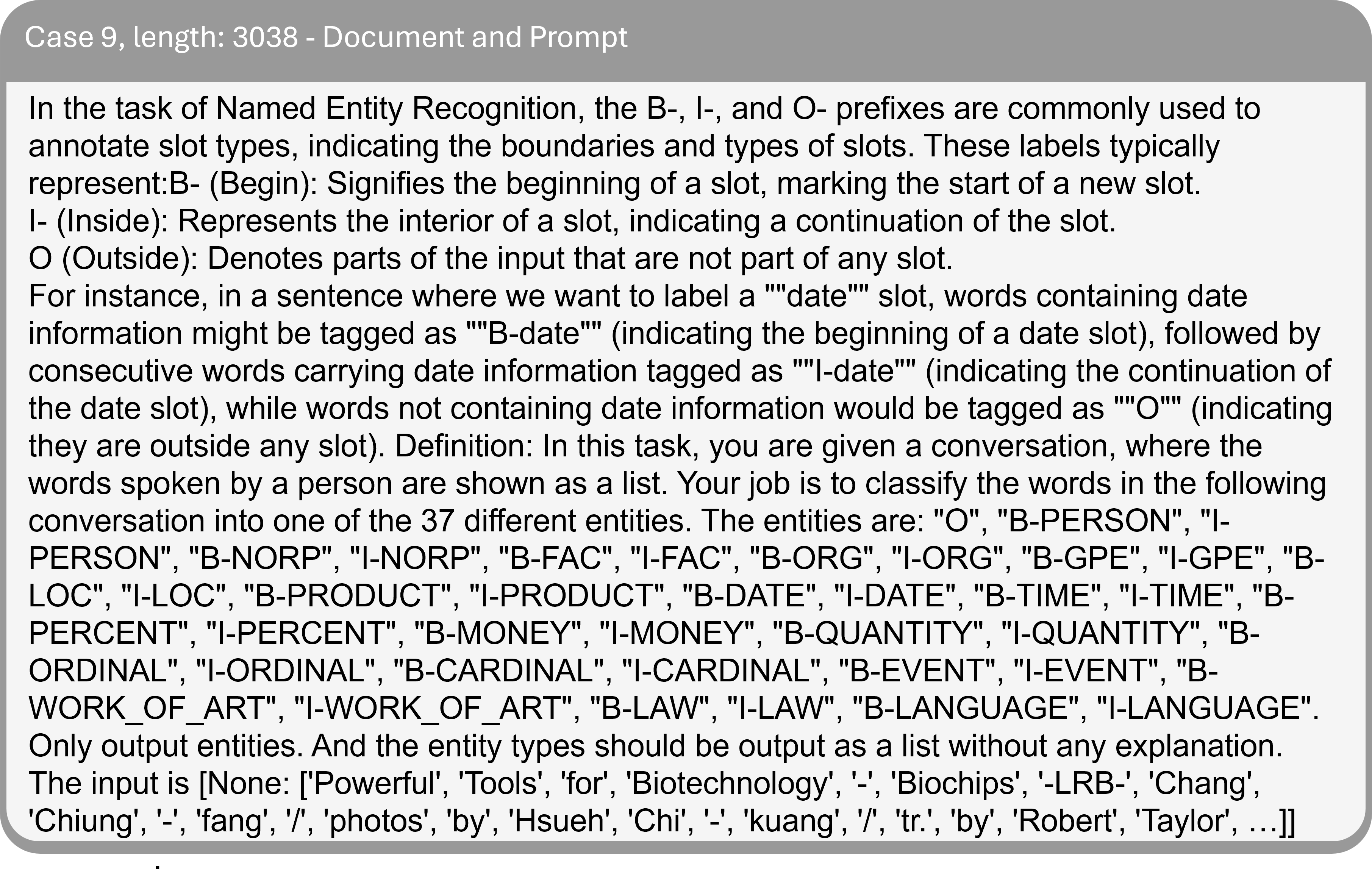} 
    \end{subfigure}
    \vspace{1pt}
    \begin{subfigure}[b]{1.0\linewidth}  
        \includegraphics[width=\textwidth]{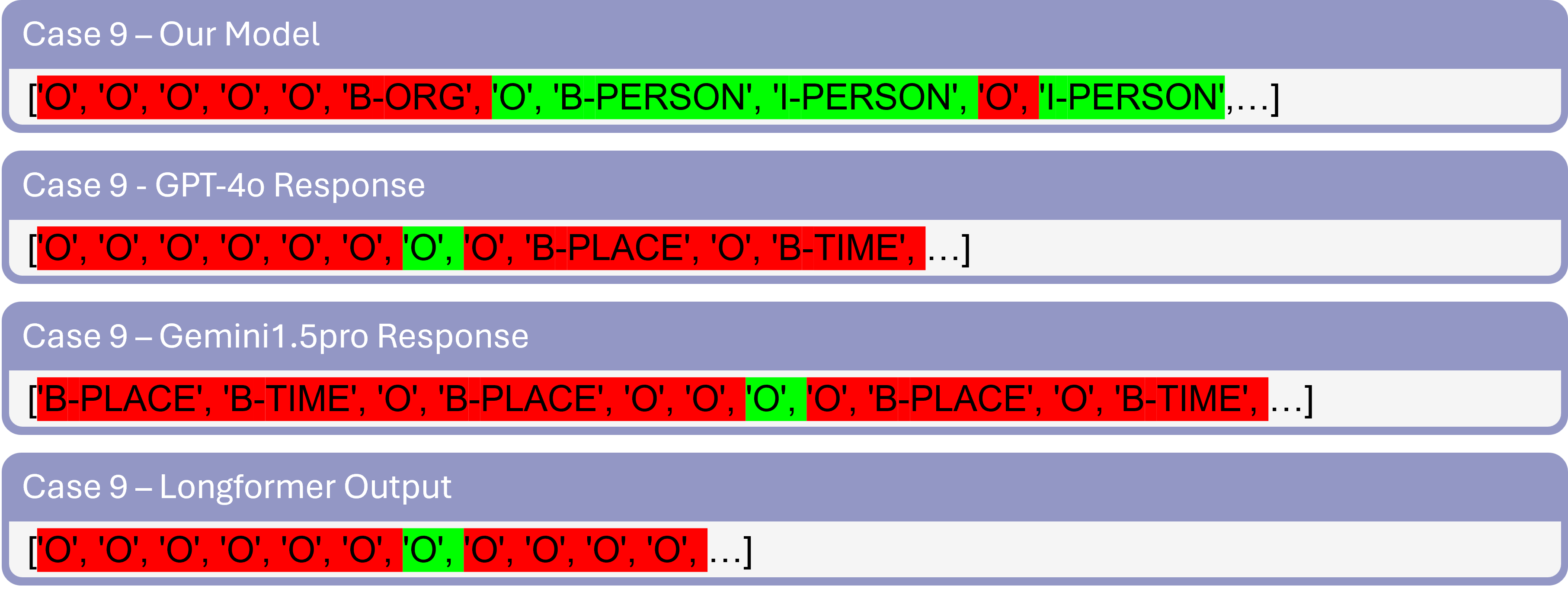}
    \end{subfigure}
    \caption{Prompt and output for a sample document of length 3038 in \textbf{CoNLL} dataset for NER task, where correct predictions are highlighted in green and wrong predictions are highlighted in red.}
    \label{case9}
\end{figure}

Table \ref{tab:Accuracy length intervals} shows that LLMs outperform Longformer in the document classification task with zero-shot prompt tuning. However, their performance drops significantly in the NER task in Table \ref{tab:Conll Microf1 length intervals} and Table \ref{tab:Gum Microf1 length intervals}. For instance, in Figure \ref{case9}, both GPT4o and Gemini1.5pro only predicted a single correct label, “O”. Moreover, the models often fail to predict a sufficient number of token labels for longer inputs, or they repeatedly predict all “O” labels or redundant label sequences. These inconsistencies suggest that LLMs struggle to generate outputs matching the input length in token classification, highlighting substantial room for improvement in this area.

\subsection{Ablation Studies}
We conducted more ablation studies, including 1) keyphrase extraction methods, 2) sentence embedding approaches, and 3) backbone model analysis, shown in Appendix \ref{app:ablation study}.

\subsection{Qualitative Analysis}
We performed a qualitative analysis by visualizing each sample document from different datasets, comparing the outputs of Longformer, GPT-4o, Gemini 1.5 Pro, and our ChuLo. ChuLo captures the context and semantic patterns of the document, providing accurate predictions, whereas the other models struggle to maintain coherence and consistency. We have more examples in Appendix \ref{app:cases}.


\section{Conclusion}
We introduced ChuLo, a novel chunk representation method that enhances the performance of Transformer-based models on long document classification and token-level classification tasks. By utilizing unsupervised keyphrase extraction, ChuLo effectively reduces input length while preserving critical information, addressing the limitations of truncation and sparse attention. Extensive experiments demonstrate that ChuLo outperforms existing methods by maintaining both global context and high accuracy, even for lengthy inputs. Our research results highlight the effectiveness of ChuLo as a robust solution for long document understanding, enabling processing of complex texts in NLP applications.

\section{Limitation}
There are several opportunities for future work, including extending the chunk representation to generative tasks such as long text generation, where chunk representation may extend the LLM's context range limitation and enhance generation quality. However, the performance of the keyphrase extraction method poses a potential risk, as its quality directly affects the overall effectiveness of the approach. We believe this work offers valuable insights into long text understanding and lays a foundation for advancements in related tasks.

\section*{Acknowledgments}
This study was supported by funding from the Google Award for Inclusion Research Program (G222897).

\bibliography{custom}

\appendix
\newpage
\section{Appendix}
\subsection{Related Works}
\label{app:related_works}

As shown in Table~\ref{tab:related_work}, most of the previous works addressing the problem of processing long documents cannot fully utilize all the content. Those methods either reduce input length via truncation or focus on local context learning to improve efficiency by applying sparse attention, approximated attention or RNN integration. Such approaches will lead to a certain level of information loss, unlike our chunking approach which can take all the content into consideration.
Hierarchical Transformer \citep{pappagari2019hierarchical} splits documents into non-overlapping chunks and computes chunk representations. RoR \citep{zhao2021ror} generates regional answers from chunks, which are combined for the final answer. However, neither considers the entire document context when chunking.
In addition, previous works applying the chunking method for processing long document context only focus on a single task, either document classification or token classification, while our framework can be applied to both tasks to guarantee both document-level and token-level understanding.

\subsection{Keyphrase Extraction}
\label{app:algorithm}
We employ the Semantic Keyphrase Prioritization (SKP) algorithm to extract keyphrases that encapsulate the key semantic information of the entire document. The detailed are shown in Algorithm \ref{alg:algorithm}.
While PromptRank uses prompts to rank keyphrases across the first segment of the document determined by its encoder model, our SKP applies this concept at the entire document level to ensure that each chunk can preserve the most informative content for the entire document. After obtaining the sorted phrases set $K_s$, we select top-n phrases as the keyphrases of the document, which can be regarded as ranked phrases according to their contextual significance within the entire document.

\begin{algorithm*}[htb]
    \scriptsize
    \caption{Semantic Keyphrase Prioritization (SKP) Algorithm}
    \label{alg:algorithm}
    \begin{multicols}{2}
    \vspace*{-7mm}
    \textbf{Input}: A tokenized document $D$, an encoder-decoder pretrained model represented by $\mathcal{F}_E$ and $\mathcal{F}_D$, a POS tagger $\mathcal{F}_{POS}$, a regular expression $\mathcal{F}_{REG} = \langle \text{NN.} \ast |\text{JJ} \rangle \ast \langle \text{NN.} \ast \rangle$\\
    \textbf{Parameter}: Experimentally determined $\alpha$, $\gamma$  \\
    \textbf{Output}: Sorted keyphrases set $K_{s}$\\
        \begin{algorithmic}[1] 
            \STATE Let $S=\emptyset$, $K_{s}=\emptyset$, $i=0$, $j=0$.
            \STATE Get the candidate phrases set: \\ $K=\mathcal{F}_{REG}(\mathcal{F}_{POS}(D))=\{k_0, k_1, \dots, k_{n-1}\}$
            \STATE Split $D$ into segments $S=\{D_0, D_1, \dots, D_{m-1}\}$ to meet the input requirement of $\mathcal{F}_E$
            \FOR{ $i < n $}
            \STATE Calculate the position penalty $r_i=\frac{L_c}{l_{d}} + \frac{\gamma}{(l_{d})^3}$\\
            where $L_c$ is the first occurrence position of $k_i$ in the document, $l_d$ is the length of the document 
            \STATE Construct the prompt $P$ “The * mainly discusses $k_i$” and tokenize, * is the category of the document.
            \FOR{ $j < m $}
            \STATE Calculate the probability $p_{ij}$ of the phrase $k_i$: \\
            $p_{ij} = \frac{1}{(l_P)^\alpha} \sum_{g=h}^{h+l_k-1} \log p(t_g \mid t_{<g})$, \\
            $p(t_g \mid t_{<g})=\mathcal{F}_D(\mathcal{F}_E(D_j), t_{<g})$ \\
            where $l_P$ is the length of the tokenized $P$, $h$ is the start index of $k_i$ in the prompt, $l_k$ is the length of $k_i$, $t$ is the token of the prompt.
            \ENDFOR
            \STATE Calculate the final score of $k_i$: $s_i=r_i\times\sum_{j=0}^{j<m}p_{ij}$
            \ENDFOR
            \STATE \textbf{return} $K_{s}=Sort(K, s)$
        \end{algorithmic}
    \end{multicols}
    \vspace*{-4mm}
\end{algorithm*}

\begin{table*}[!ht]
    \centering
    \scriptsize
    \setlength{\tabcolsep}{13pt}
    \begin{tabular}{l|cccc}
    \hline
        \textbf{Model} & \textbf{Year} & \textbf{Task} & \textbf{Lengthy Document Solution} & \textbf{Core Architecture} \\ \hline
        Efficient Classification \citep{park2022efficient} & 2022 & D & Truncating & Transformer  \\ 
        Hierarchical transformer \citep{pappagari2019hierarchical} & 2019 & D & Chunking (Partial, Phrase)& Transformer  \\ 
        RoR \citep{zhao2021ror} & 2021 & T & Chunking (Partial, Voting) & Transformer  \\ 
        Longformer \citep{beltagy2020longformer} & 2020 & D, T & Sparse Attention & Transformer  \\ 
        BigBird \citep{zaheer2020big} & 2020 & D, T & Sparse Attention & Transformer  \\ 
        Routing Transformer \citep{roy2021efficient} & 2021 & D, T, G & Sparse Attention & Transformer  \\ 
        Macformer \citep{peng2021random} & 2021 & D, T & Approximated Attention & Transformer  \\ 
        Linformer \citep{wang2020linformer} & 2020 & D, T, G & Approximated Attention & Transformer  \\ 
        Performer \citep{choromanski2020masked} & 2020 & D, T, G & Approximated Attention & Transformer  \\ 
        Transformer-xl \citep{dai2019transformer} & 2019 & G & RNN Integration & Transformer  \\ 
        Block-Recurrent Transformer \citep{hutchins2022block} & 2022 & G & RNN Integration & Transformer  \\ 
        RAN \citep{li-etal-2023-recurrent} & 2023 & D, T & RNN Integration & Attention  \\ 
        \citep{ccetindaug2023named} & 2023 & T & N/A & LSTM  \\ 
        \citep{mengliev2024developing} & 2024 & T & N/A & Neural Network  \\ 
        \citep{park2023web} & 2023 & T & N/A & Transformer  \\ 
        \citep{bhattacharya2023improving} & 2023 & T & N/A & LSTM  \\ 
        Gpt-NER \citep{wang2023gpt} & 2023 & T & N/A & Transformer  \\ 
        \citep{dagdelen2024structured} & 2024 & T & N/A & Transformer  \\ 
        \citep{hu2024improving} & 2024 & T & N/A & Transformer  \\ 
        \citep{yu2023grounded} & 2023 & T & N/A & Transformer  \\ 
        \citep{zhang2023reducing} & 2023 & T & N/A & Transformer  \\ 
        \hline
        \textbf{Ours} & \textbf{2024} & \textbf{D, T} & \textbf{Chunking (Entire)} & \textbf{Transformer} \\ \hline
    \end{tabular}
    \caption{Summary of Related Works. D, T, G represent tasks of document classification, token classification, and text generation, respectively. }
    \label{tab:related_work}
\end{table*}

\subsection{Baselines}
\label{app:baseline models}

We use BERT \citep{kenton2019bert} as our backbone model, comparing it with ToBERT \citep{pappagari2019hierarchical}, CogLTX \citep{ding2020cogltx}, Longformer \citep{beltagy2020longformer}, various BERT variants \citep{park2022efficient} and ChunkBERT \citep{jaiswal2023breaking} for the document classification task. For the NER task, we compare against Longformer, BigBird \citep{zaheer2020big}, and two large language models, GPT4o and Gemini1.5pro. Below are brief descriptions of the baseline models:

\begin{itemize}
 \item\textbf{BERT}: A transformer model pre-trained on masked language modeling (MLM) and next-sentence prediction (NSP). We fine-tuned the BERT-base variant on each dataset. 

 \item\textbf{ToBERT}: A BERT variant designed for long document classification, utilizing an additional transformer layer to learn inter-chunk relationships.

 \item\textbf{CogLTX}: A framework for applying BERT to long documents by training a key sentence identification model to assist in document understanding

 \item\textbf{Longformer}: Optimized for long sequences using sparse attention, combining dilated sliding window and global attention patterns

 \item\textbf{BigBird}: Utilizes block sparse attention, integrating sliding window, global, and random attention patterns across token blocks.

 \item\textbf{BERT+TextRank} and \textbf{BERT+Random}: Proposed to select other tokens randomly or with the help of TextRank\citep{mihalcea2004textrank} to feed into the BERT model as the supplementation for long document classification.

 \item\textbf{ChunkBERT}: A BERT variant for long document classification that processes self-attention within document chunks and adds a TextCNN module for classification using the chunk representation.

 \item\textbf{GPT-4o}: A transformer-based multi-modal large language model developed by OpenAI, which leverages large-scale pretraining data to process diverse language tasks via instruction prompts.

 \item \textbf{Gemini 1.5 Pro}: an advanced multi-modal AI model from Google, leveraging a Sparse Mixture-of-Experts (MoE) Transformer architecture, with a context window of up to 2 million tokens. This architecture allows for the efficient handling of long documents.
 
\end{itemize}

\subsection{Baseline Input}
\label{app: baseline input}
We selected these baseline models because they represent diverse methods for processing long documents. As summarized in Table \ref{tab:baseline_input}, BERT truncates the input to 512 tokens. Longformer and BigBird utilize sparse attention mechanisms, allowing them to process up to 4096 tokens while conserving computational resources. ToBERT divides the input into 200-token chunks, feeds them to BERT for chunk representations, and uses a transformer layer for downstream tasks. However, it cannot capture dependencies across the entire input sequence. CogLTX selects key sentences to form a 512-token input, limiting its input size to that constraint. BERT variants like BERT+TextRank and BERT+Random select up to 512 tokens using TextRank or random selection. They concatenate the [CLS] representation of the initial 512 tokens with the selected tokens, creating an augmented input for a fully connected classification layer, with a maximum input length of 1024 tokens. ChunkBERT splits the input into chunks, computes self-attention, and feeds chunk representations into a TextCNN module for classification. The original implementation processes up to 4096 tokens per document. It has the same limitation as the ToBERT. For GPT4o and Gemini1.5pro, we input all tokens together with our instruction in the prompt due to the large input size supported by these large language model. In contrast to these baseline models, our approach flexibly segments the entire input into chunks of varying sizes, using semantic keyphrases to minimize information loss. Additionally, we compute chunk-level attention to capture long-range dependencies more effectively.

\begin{table*}[t]
\setlength\tabcolsep{10pt}
\scriptsize
\centering
\begin{tabular}{l|c}
\noalign{\hrule height 0.8pt}
\textbf{Model} & \textbf{Input} \\
\hline
BERT \citep{kenton2019bert} & The first 512 tokens \\
ToBERT \citep{pappagari2019hierarchical} & Segmented all input tokens\\
CogLTX \citep{ding2020cogltx} & Selected 512 tokens \\
Longformer \citep{beltagy2020longformer} & The first 4096 tokens \\
BigBird \citep{zaheer2020big} & The first 4096 tokens \\
BERT+TextRank \citep{park2022efficient} & The first 512 tokens with the selected 512 tokens\\
BERT+Random \citep{park2022efficient}  & The first 512 tokens with the selected 512 tokens \\
ChunkBERT \citep{jaiswal2023breaking}  & The first 4096 tokens \\
GPT4o & All input tokens with instruction \\
Gemini1.5pro & All input tokens with instruction \\

\noalign{\hrule height 0.8pt}
\end{tabular}
\caption{The input of the baseline models}
\label{tab:baseline_input}
\end{table*}

\subsection{Details of datasets}
\label{sec:Appendix:Dataset statistics}
\begin{table}[H]
\scriptsize
\centering
\begin{tabular}{l|ccc}
\noalign{\hrule height 0.8pt}
\multicolumn{1}{l|}{\textbf{Datasets}}  & \textbf{Train/Dev/Test} & \textbf{\#Classes} & \textbf{Avg. Length} \\
\hline
HP & 516/64/65 & 2 & 705 \\
LUN & 12003/2992/2250 & 3 & 480 \\
EURLEX57k & 45000/6000/6000 & 4271 & 707 \\
-INVERTED & 45000/6000/6000 & 4271 & 707 \\
\hline
GUM & 179/26/26 & 21 & 972 \\
CoNLL & 120/20/20 & 37 & 5065 \\
\noalign{\hrule height 0.8pt}
\end{tabular}
\caption{The split and statistics of the datasets, including document classification task (HP, LUN, EURLEX57K, and Inverted EURLEX57K) and token classification task (GUM, CoNLL)}
\label{datasettable}
\end{table}
We analyzed the data distribution across the datasets used in this paper. Of these, the CoNLL dataset has the highest average number of tokens per document at 5,065. In contrast, LUN has the shortest average length, with 480 tokens per document. Both HP and EURLEX57K have similar average document lengths, measuring 705 and 707 tokens respectively. GUM presents a relatively higher token count, averaging 972 tokens per document.

Regarding the number of classes, EURLEX57K is the most complex dataset, containing 4,271 unique labels. In comparison, HP and LUN are more limited, with only 2 and 3 classes respectively. GUM and CoNLL are more diverse, with 21 and 37 different classes. Beyond label diversity, EURLEX57K also has the largest sample size, comprising 45,000 training samples, 6,000 development samples, and 6,000 test samples. LUN is the second-largest dataset, with 12,003 training samples, 2,992 development samples, and 2,250 test samples. Due to our selection strategy, CoNLL has the longest average document length, with the fewest samples. It has a total of 160 documents split into 120/20/20 for training, development, and test sets. GUM follows a similar distribution with 179/26/26 samples. The HP dataset includes 516 training samples, 64 development samples, and 65 test samples.

\subsection{Implementation details}
\label{app:imp. details}

\subsubsection{Experiment hyperparameters}
 We performed extensive experiments to select the hyperparameters, including chunk size, token weights, learning rates, and warm-up strategies and steps. The optimal hyperparameters for each dataset for our proposed ChuLo model are presented in Table \ref{tap:app_hp_imp_details}. 

\begin{table*}[htp]
\centering
\setlength\tabcolsep{12.5pt}
\scriptsize
    \begin{tabular}{l|c|c|c|c|c|c}
        \noalign{\hrule height 0.8pt}
        \multicolumn{1}{l|}{\textbf{Hyperparameter}} & \textbf{HP} & \textbf{LUN} & \textbf{EURLEX57K} & \textbf{I-EURLEX57K} & \textbf{CoNLL} & \textbf{GUM}\\
        \hline
        Number of top-n phrases & 15 & 15 & 15 & 15 & 15 & 15 \\
        \hline
        Chunk size $n$ & 10 & 50 & 5 & 5 & 20 & 50 \\
        \hline
        Weight for $T_k$ & 0.8 & 0.5 & 0.8 & 0.8 & 0.8 & 0.8 \\
        \hline
        Weight for $T_{nk}$ & 0.1 & 0.1 & 0.1 & 0.1 & 0.1 & 0.1 \\
        \hline
        Learning Rate & 5e-5 & 5e-5 & 5e-5 & 5e-5 & 5e-5 & 5e-5 \\
        \hline
        Batch Size & 16 & 32 & 16 & 16 & 2 & 8  \\
        \hline
        Warm-up Strategy & Linear & Linear & Cosine & Cosine & Linear & Linear \\
        \hline
        Warm-up Steps & 10\% & 10\% & 5\% & 5\% & 10\% & 10\% \\
        \hline
        Mex epoch & 100 & 100 & 100 & 100 & 100 & 100 \\
        \hline
        Stop Patience & 10 & 10 & 10 & 10 & 10 & 10 \\
        \hline
        Optimizer  & AdamW & AdamW & AdamW & AdamW & AdamW & AdamW \\
        \hline
        Optimizer Weight Decay & 1e-2 & 1e-2 & 1e-2 & 1e-2 & 1e-2 & 1e-2  \\
        \hline
        Optimizer Betas & 0.9, 0.999 & 0.9, 0.999 & 0.9, 0.999 & 0.9, 0.999 & 0.9, 0.999 & 0.9, 0.999 \\
        \noalign{\hrule height 0.8pt}
    \end{tabular}
    \caption{The optimal hyperparameters used in our experiments.}
    \label{tap:app_hp_imp_details}
\end{table*}



\subsubsection{Hardware Information}
Our experiments are run on the Linux platform with an A6000 Nvidia graphic card and an AMD Ryzen Threadripper PRO 5955WX 16-core CPU, and the RAM is 128G.

\subsection{Ablation Studies}
\label{app:ablation study}

We performed a few ablation studies on the HP and LUN  to assess the impact of various components within our model. First, we analyzed the effectiveness of different keyphrase extraction methods and the effect of using average chunk representations. As shown in Table \ref{tab:keyphrase ablation}, the PromptRank-based method yields the highest performance across both datasets, outperforming alternatives like YAKE-based. This improvement can be attributed to PromptRank’s ability to extract higher-quality keyphrases by considering semantic relationships within the document, whereas YAKE relies primarily on statistical features such as phrase frequency, resulting in less semantically rich keyphrases. 
Then, we explored the effect of incorporating sentence embeddings into the chunk representations to introduce global sentence-level context. Surprisingly, as shown in Table \ref{tab:sent emb ablation}, the results indicate a performance drop when sentence embeddings are included. We hypothesize that adding sentence-level information at the initial representation stage may cause chunk embeddings within the same sentence to become too similar, hindering the model’s ability to learn distinctive patterns and reducing overall classification performance.
Then, we evaluated the performance of different backbone models for the chunk attention module while keeping the keyphrase extraction and chunk representation settings consistent. Table \ref{tab:backbone ablation} shows that BERT outperforms Longformer as the backbone. This result suggests that, after document chunking, the input sequences become relatively short, making it difficult for Longformer to leverage its long-range attention capabilities fully. Consequently, Longformer may suffer underutilization during training, resulting in suboptimal performance compared to BERT. 


\begin{table}[ht]
    \setlength{\tabcolsep}{18pt}
    \scriptsize
    \centering
    \begin{tabular}{p{0.3\linewidth}|p{0.1\linewidth}p{0.1\linewidth}l|cc}
    \noalign{\hrule height 0.8pt}
    \textbf{Keyphrase method} & \textbf{HP} & \textbf{LUN}\\
    \hline
    Average   & \textbf{0.9538} & 0.5951\\
    YAKE   & 0.8769 & 0.5951\\
    PromptRank   & \textbf{0.9538} & \textbf{0.6440}\\
    \noalign{\hrule height 0.8pt}
    \end{tabular}
    \caption{Effect of keyphrase extraction methods; Average: Average Chunk Representations}
    \label{tab:keyphrase ablation}
    \vspace{-20pt}
\end{table}

\begin{table}[htp]
    \scriptsize
    \centering
    \setlength{\tabcolsep}{18pt}
    \begin{tabular}{p{0.3\linewidth}|p{0.1\linewidth}p{0.1\linewidth}}
    \noalign{\hrule height 0.8pt}
    \textbf{Sentence Embedding} & \textbf{HP} & \textbf{LUN}\\
    \hline
    w/o sentence emb.   & \textbf{0.9538} & \textbf{0.6440}\\
    sentence emb.   & 0.9076 & 0.5537\\
    \noalign{\hrule height 0.8pt}
    \end{tabular}
    \caption{Effect of sentence embedding, adding the sentence-level information to the chunk representations.}
    \label{tab:sent emb ablation}
\end{table}

\begin{table}[ht]
    \scriptsize
    \centering
    \vspace{-20pt}
    \setlength{\tabcolsep}{18pt}
    \begin{tabular}{p{0.3\linewidth}|p{0.1\linewidth}p{0.1\linewidth}l|cc}
    \noalign{\hrule height 0.8pt}
    \textbf{Backbone} & \textbf{HP} & \textbf{LUN}\\
    \hline
    BERT (Ours) & \textbf{0.9538} & \textbf{0.6440}\\
    RoBERTa   & 0.8615 & 0.5906\\
    Longformer & 0.8923 & 0.5600\\
    \noalign{\hrule height 0.8pt}
    \end{tabular}
    \caption{Effect of different backbone models for the chunk attention. }
    \label{tab:backbone ablation}
\vspace{-10pt}
\end{table}

\subsection{More Case Studies}
\label{app:cases}

\begin{figure}[!h]
    \vspace{-5pt}
    \centering
    \includegraphics[width=1.0\linewidth]{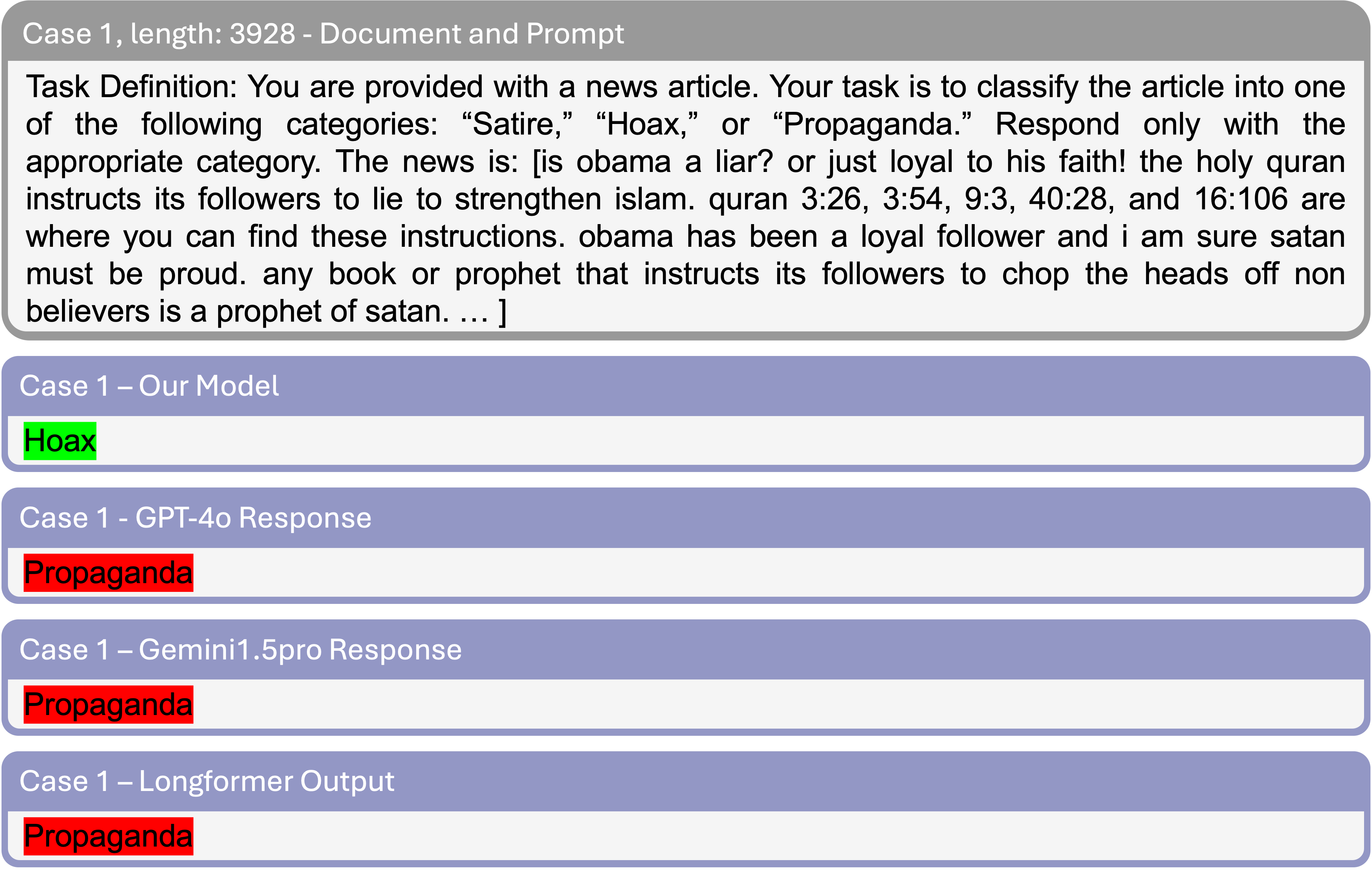}
    \caption{Prompt and output for a sample document of length 3928 in \textbf{LUN} dataset, where the correct prediction is highlighted in green and wrong predictions are highlighted in red. Compared to GPT4o, Gemini1.5pro and Longformer, our model can \textbf{correctly} classify the given document as \textbf{Hoax}.}
    \label{fig:case1}
\end{figure}

\begin{figure}[!h]
    \centering
    \includegraphics[width=1.0\linewidth]{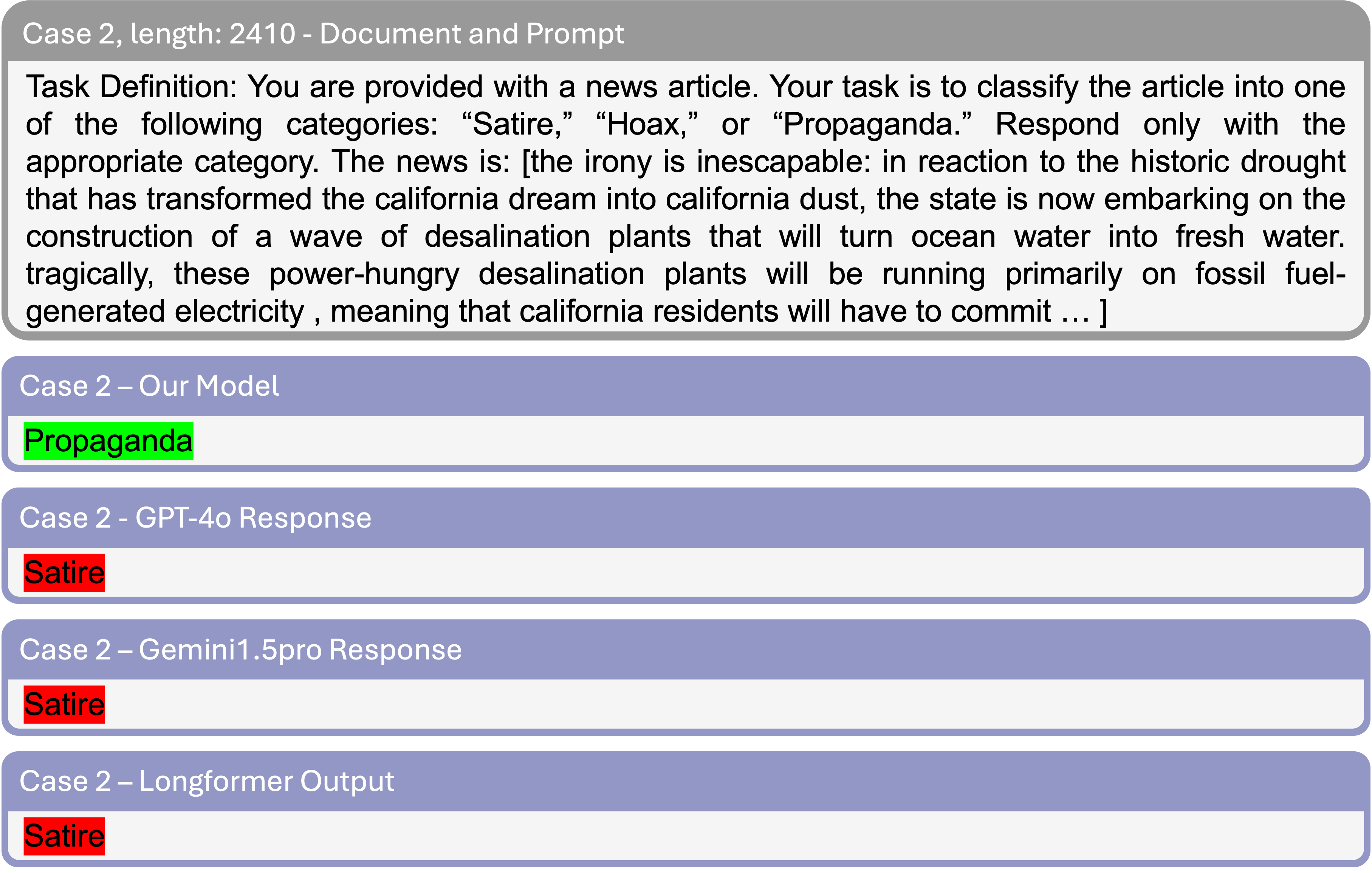}
    \caption{Prompt and output for a sample document of length 2410 in \textbf{LUN} dataset, where the correct prediction is highlighted in green and wrong predictions are highlighted in red. Compared to GPT4o, Gemini1.5pro and Longformer, our model can \textbf{correctly} classify the given document as \textbf{Propaganda}.}
    \label{fig:case2}
\end{figure}

\begin{figure}[!h]
    \vspace{-5pt}
    \centering
    \includegraphics[width=1.0\linewidth]{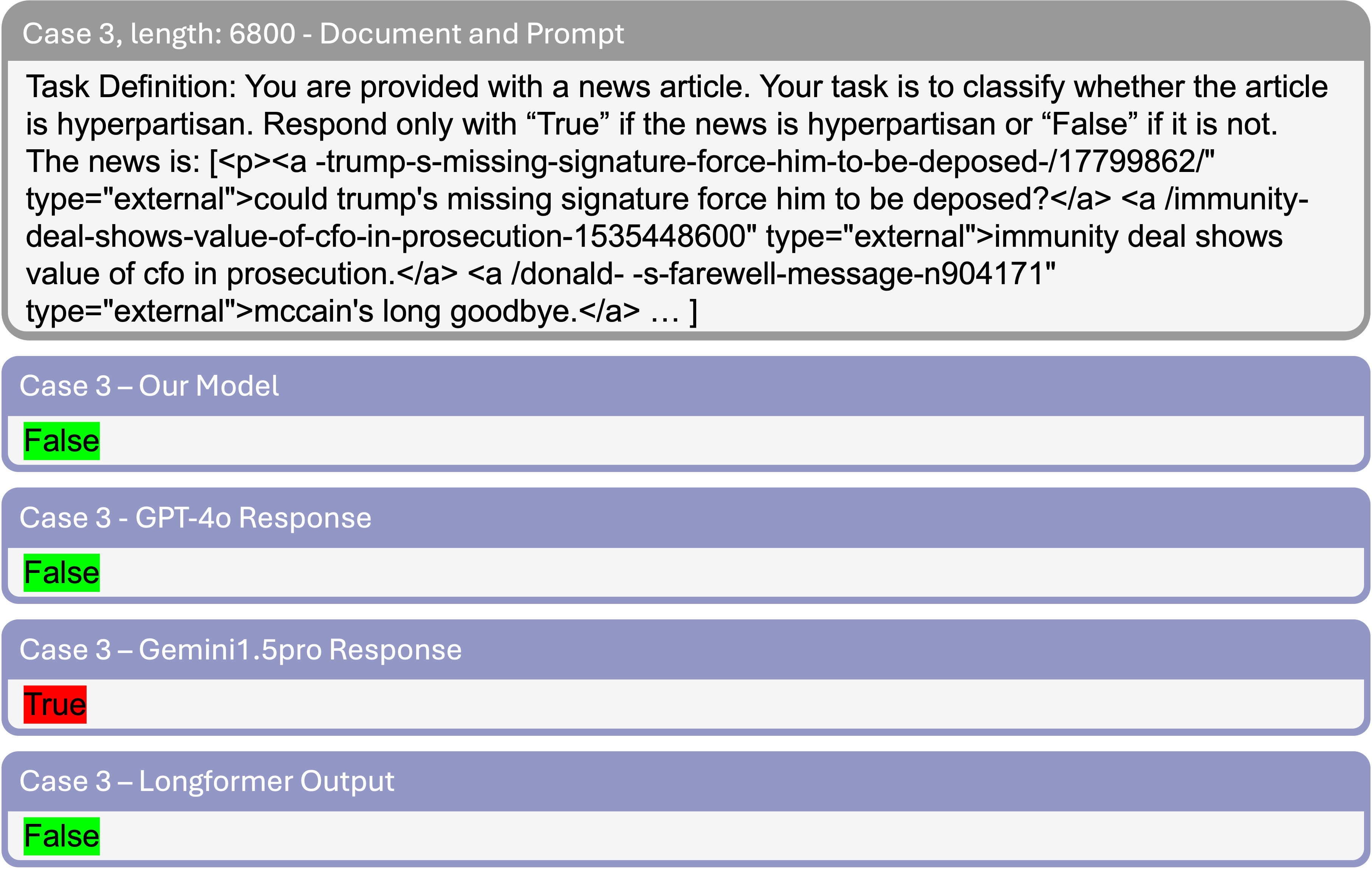}
    \caption{Prompt and output for a sample document of length 6800 in \textbf{Hyperpartisan} dataset, where correct predictions are highlighted in green and the wrong prediction is highlighted in red. Compared to Gemini1.5pro, our model, GPT4o and Longformer can \textbf{correctly} classify the given document as \textbf{False}.}
    \label{fig:case3}
\end{figure}

\begin{figure}[!h]
    \centering
    \includegraphics[width=1.0\linewidth]{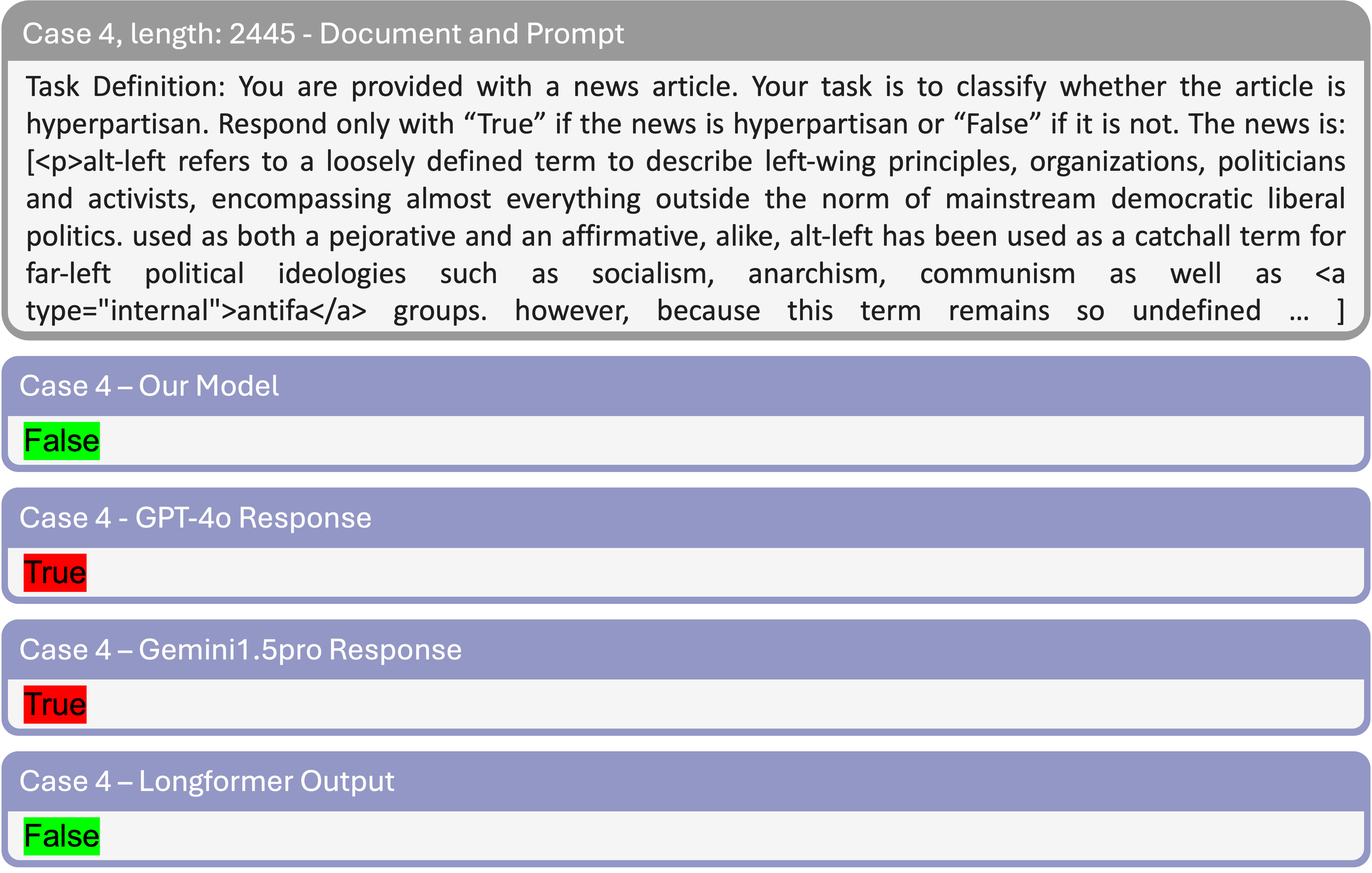}
    \caption{Prompt and output for a sample document of length 2445 in \textbf{Hyperpartisan} dataset, where correct predictions are highlighted in green and wrong predictions are highlighted in red. Compared to GPT4o and Gemini1.5pro, our model and Longformer can \textbf{correctly} classify the given document as \textbf{False}.}
    \label{fig:case4}
\end{figure}

\begin{figure}[!h]
    \centering
    \begin{subfigure}[b]{1.0\linewidth}   
        \includegraphics[width=\textwidth]{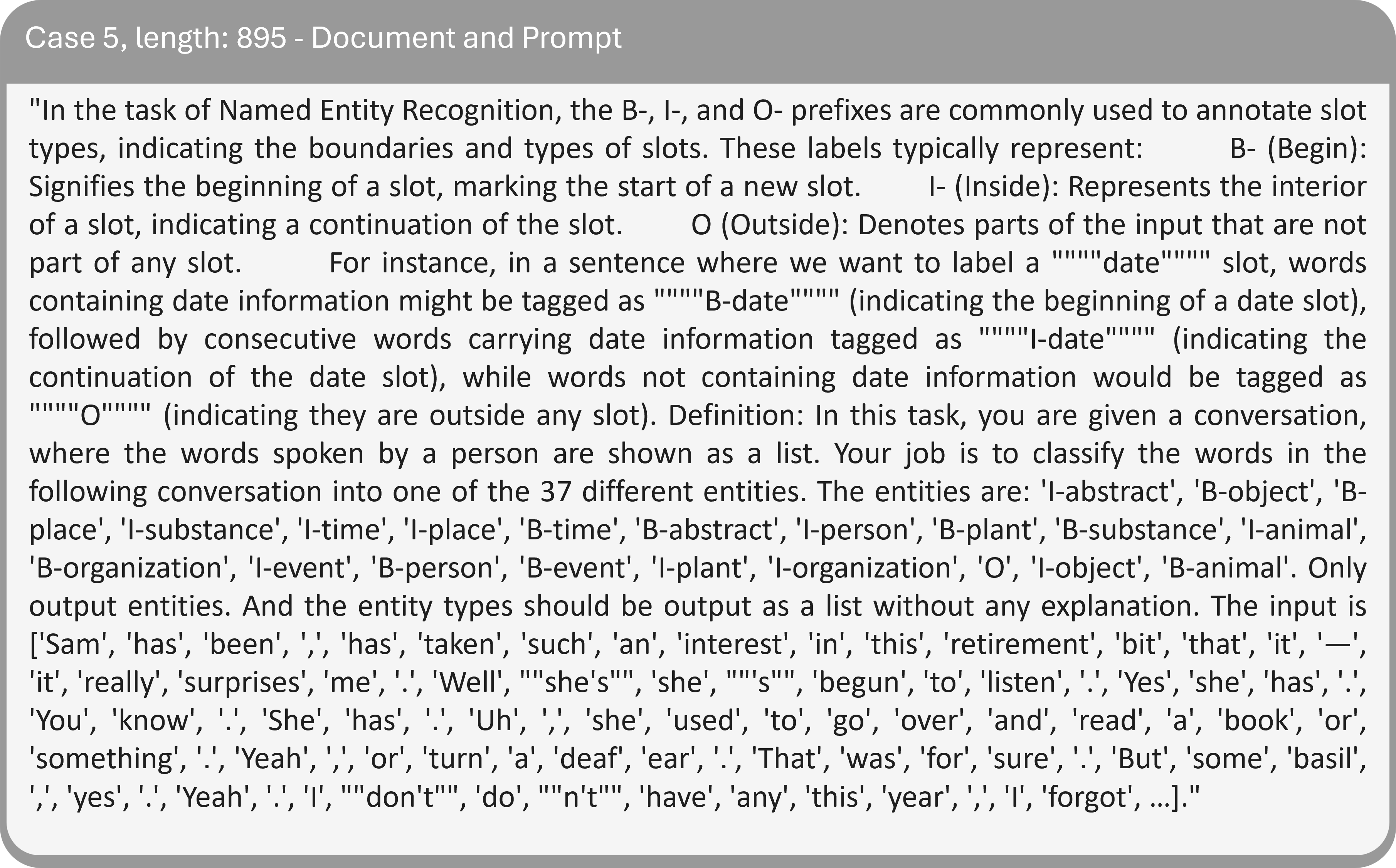} 
    \end{subfigure}
    \vspace{1pt}
    \begin{subfigure}[b]{1.0\linewidth}  
        \includegraphics[width=\textwidth]{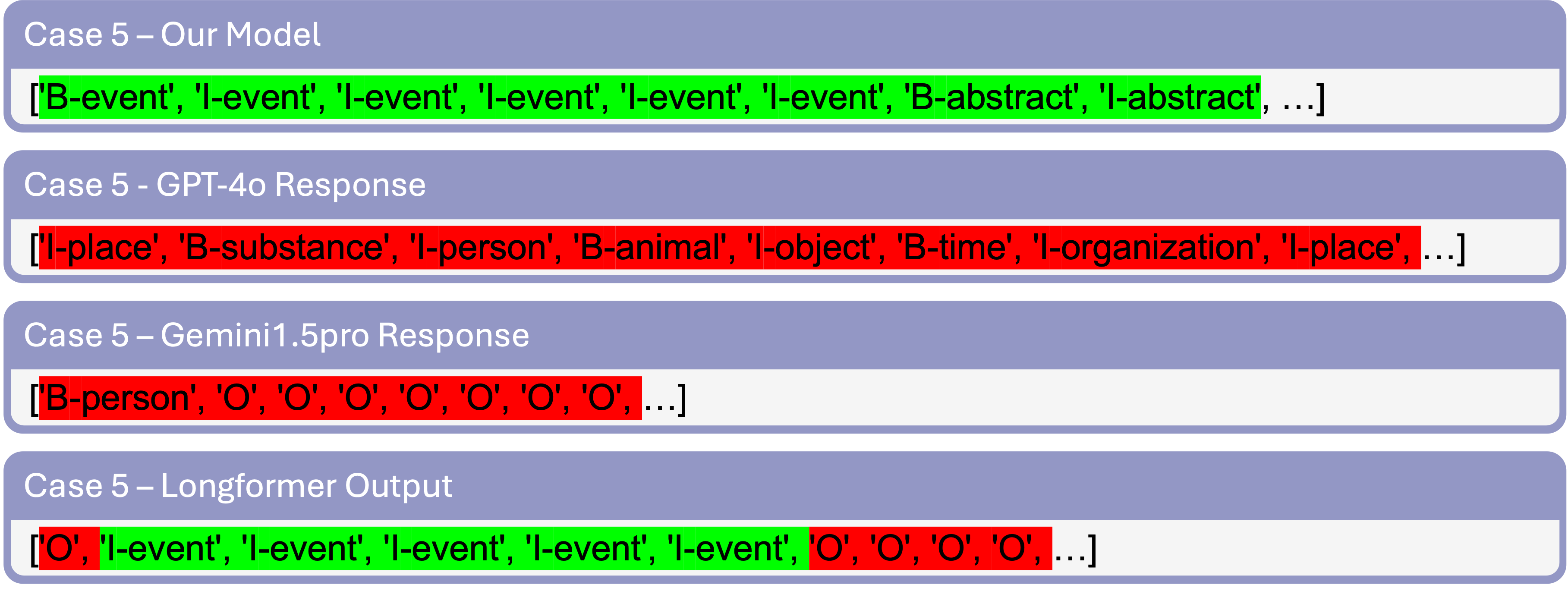}
    \end{subfigure}
    \caption{Prompt and output for a sample document of length 895 in \textbf{GUM} dataset for NER task, where correct predictions are highlighted in green and wrong predictions are highlighted in red.}
    \label{case5}
\end{figure}

\begin{figure}[!h]
    \centering
    \begin{subfigure}[b]{1.0\linewidth}   
        \includegraphics[width=\textwidth]{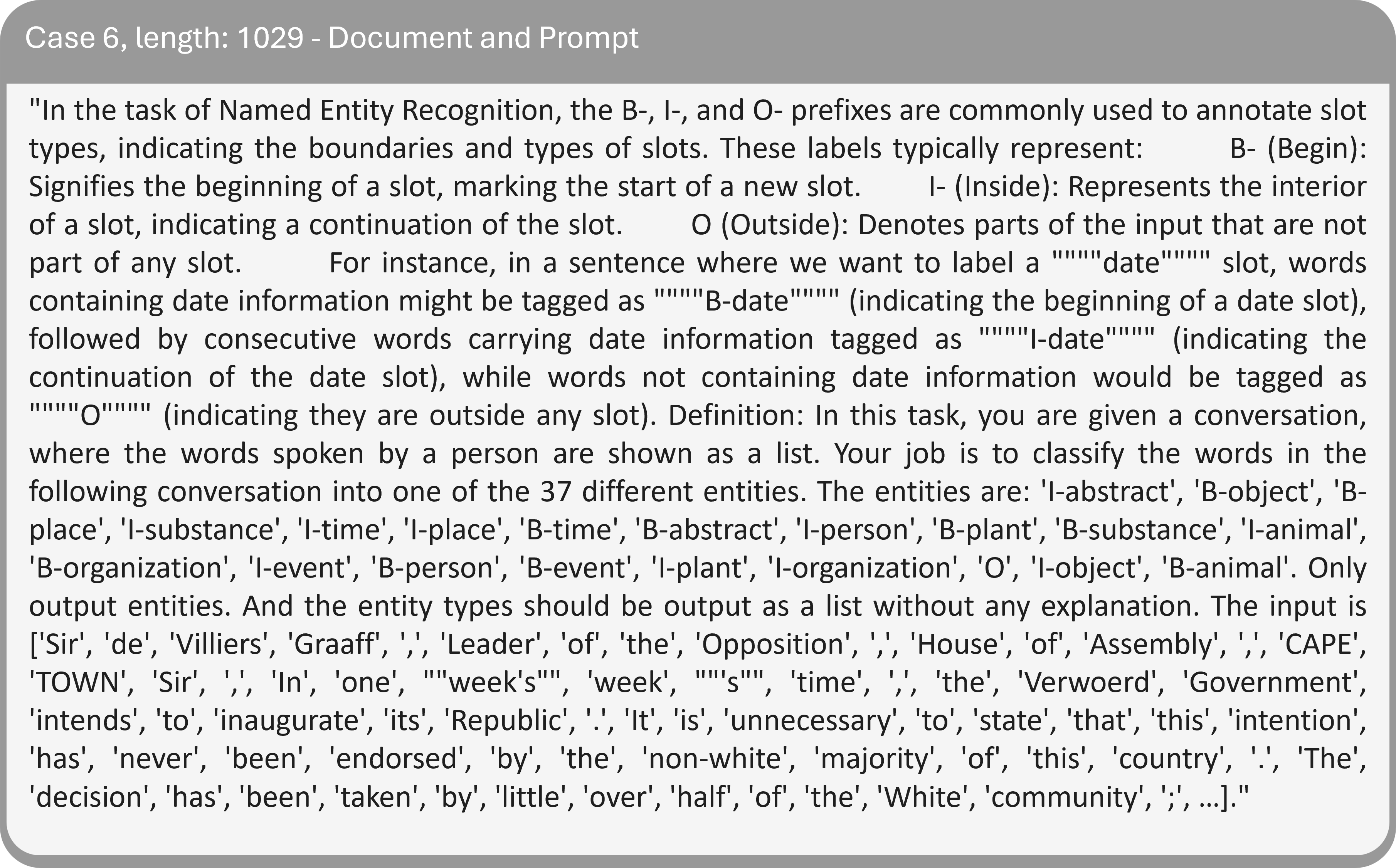} 
    \end{subfigure}
    \vspace{1pt}
    \begin{subfigure}[b]{1.0\linewidth}  
        \includegraphics[width=\textwidth]{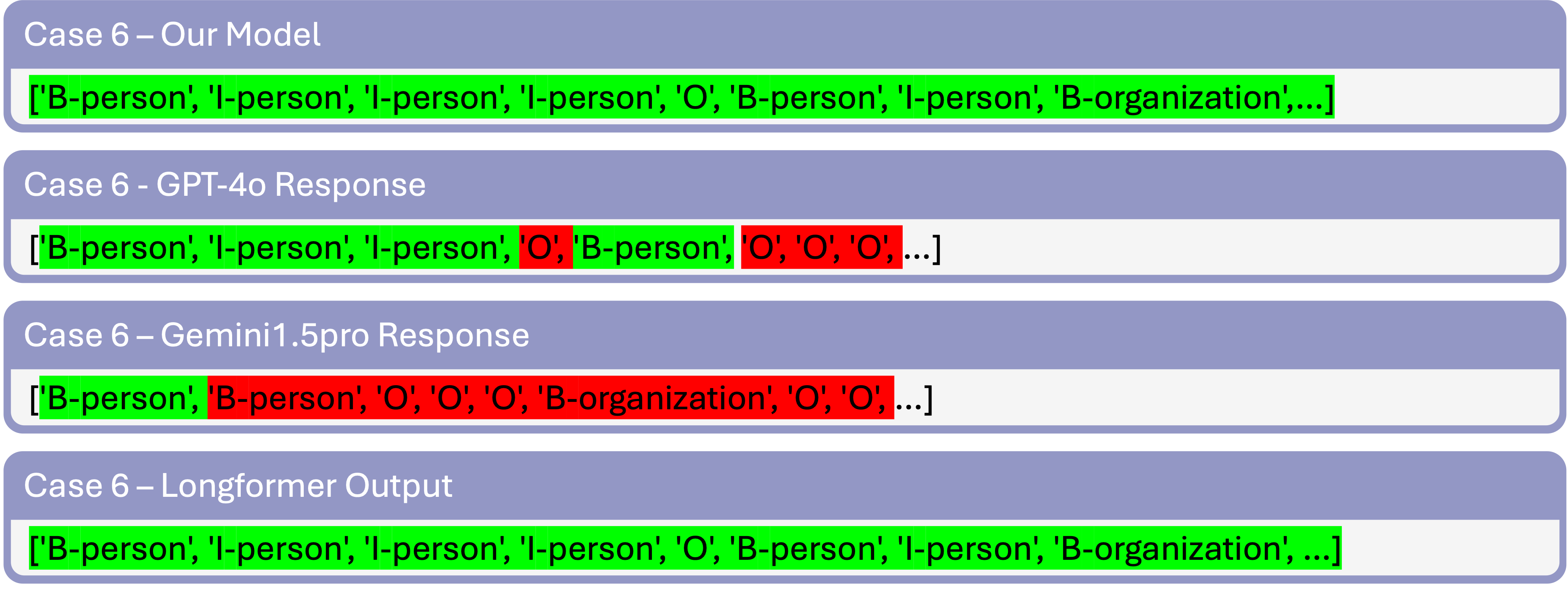}
    \end{subfigure}
    \caption{Prompt and output for a sample document of length 1029 in \textbf{GUM} dataset for NER task, where correct predictions are highlighted in green and wrong predictions are highlighted in red.}
    \label{case6}
\end{figure}

\begin{figure}[!h]
    \centering
    \begin{subfigure}[b]{1.0\linewidth}   
        \includegraphics[width=\textwidth]{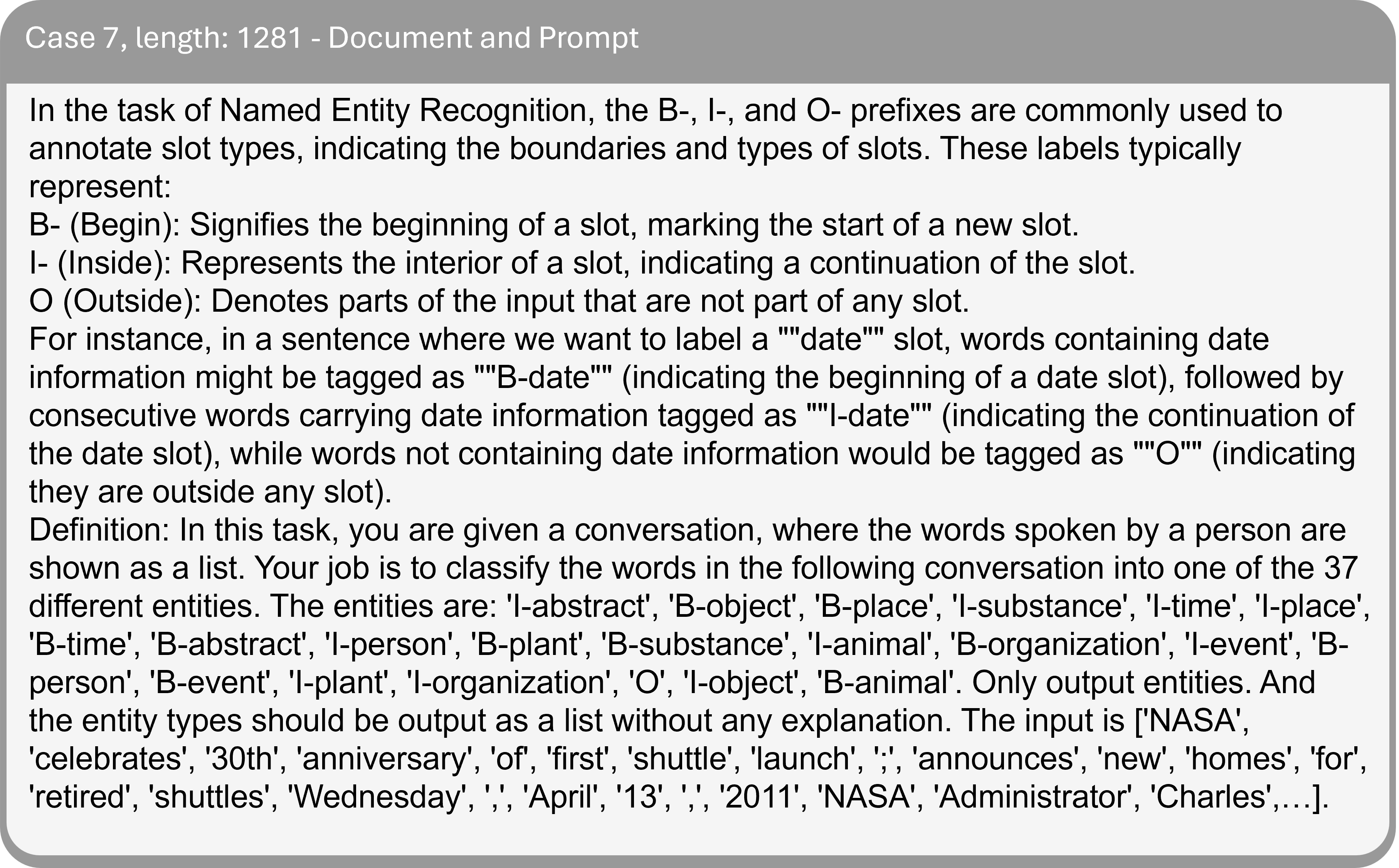} 
    \end{subfigure}
    \vspace{1pt}
    \begin{subfigure}[b]{1.0\linewidth}  
        \includegraphics[width=\textwidth]{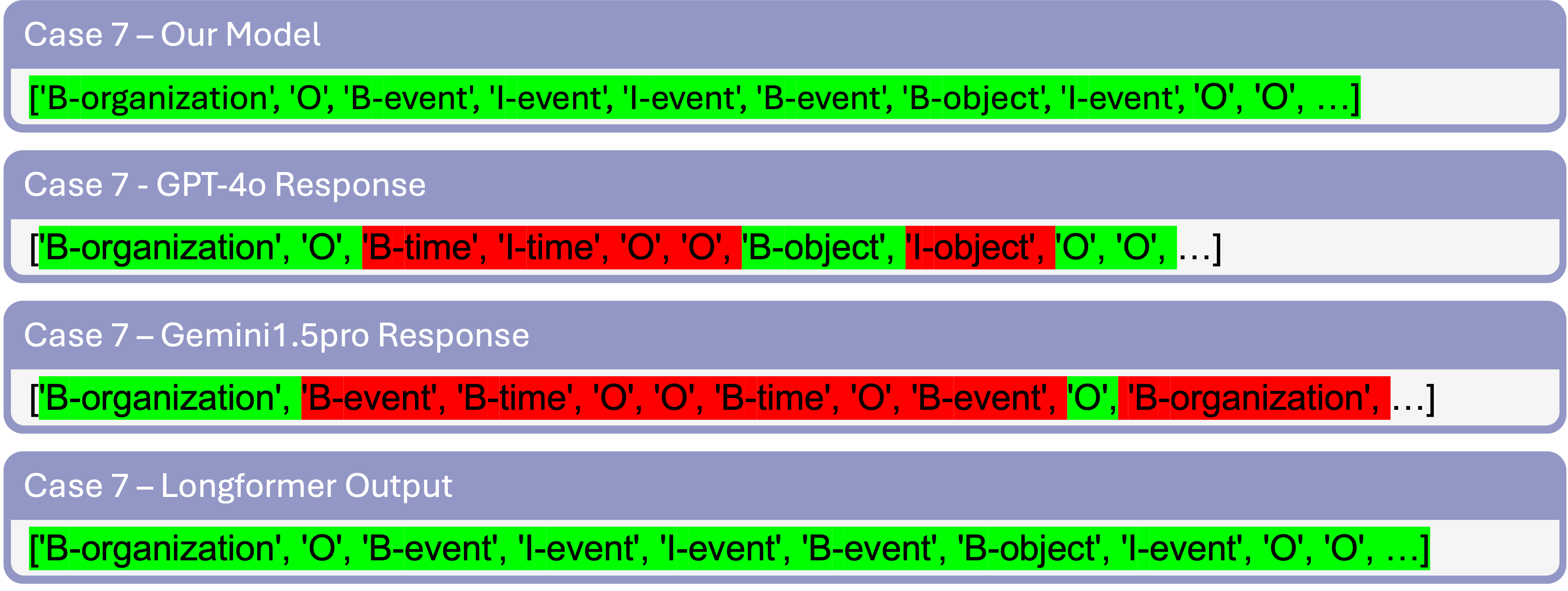}
    \end{subfigure}
    \caption{Prompt and output for a sample document of length 1281 in \textbf{GUM} dataset for NER task, where correct predictions are highlighted in green and wrong predictions are highlighted in red.}
    \label{case7}
\end{figure}

\begin{figure}[!h]
    \centering
    \begin{subfigure}[b]{1.0\linewidth}   
        \includegraphics[width=\textwidth]{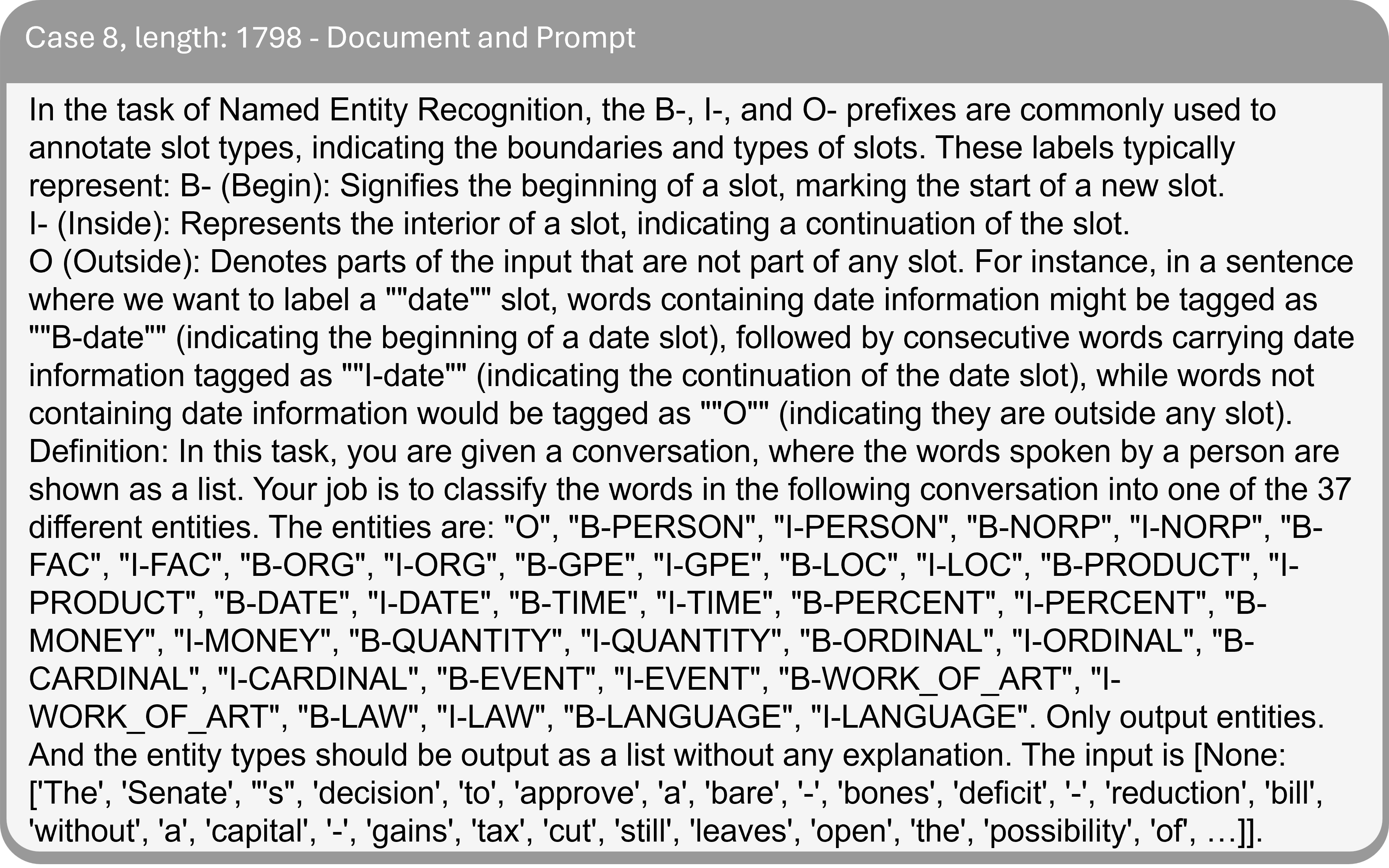} 
    \end{subfigure}
    \vspace{1pt}
    \begin{subfigure}[b]{1.0\linewidth}  
        \includegraphics[width=\textwidth]{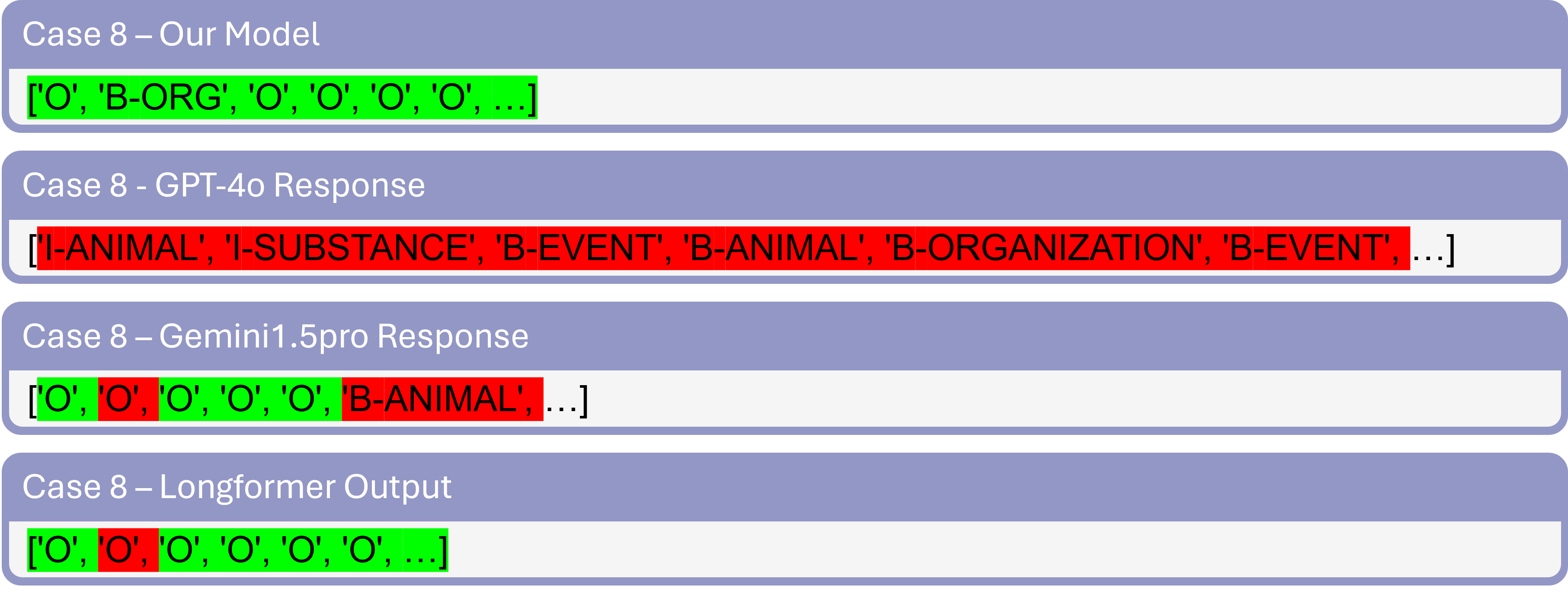}
    \end{subfigure}
    \caption{Prompt and output for a sample document of length 1798 in \textbf{CoNLL} dataset for NER task, where correct predictions are highlighted in green and wrong predictions are highlighted in red.}
    \label{case8}
\end{figure}

\begin{figure}[!h]
    \centering
    \begin{subfigure}[b]{1.0\linewidth}   
        \includegraphics[width=\textwidth]{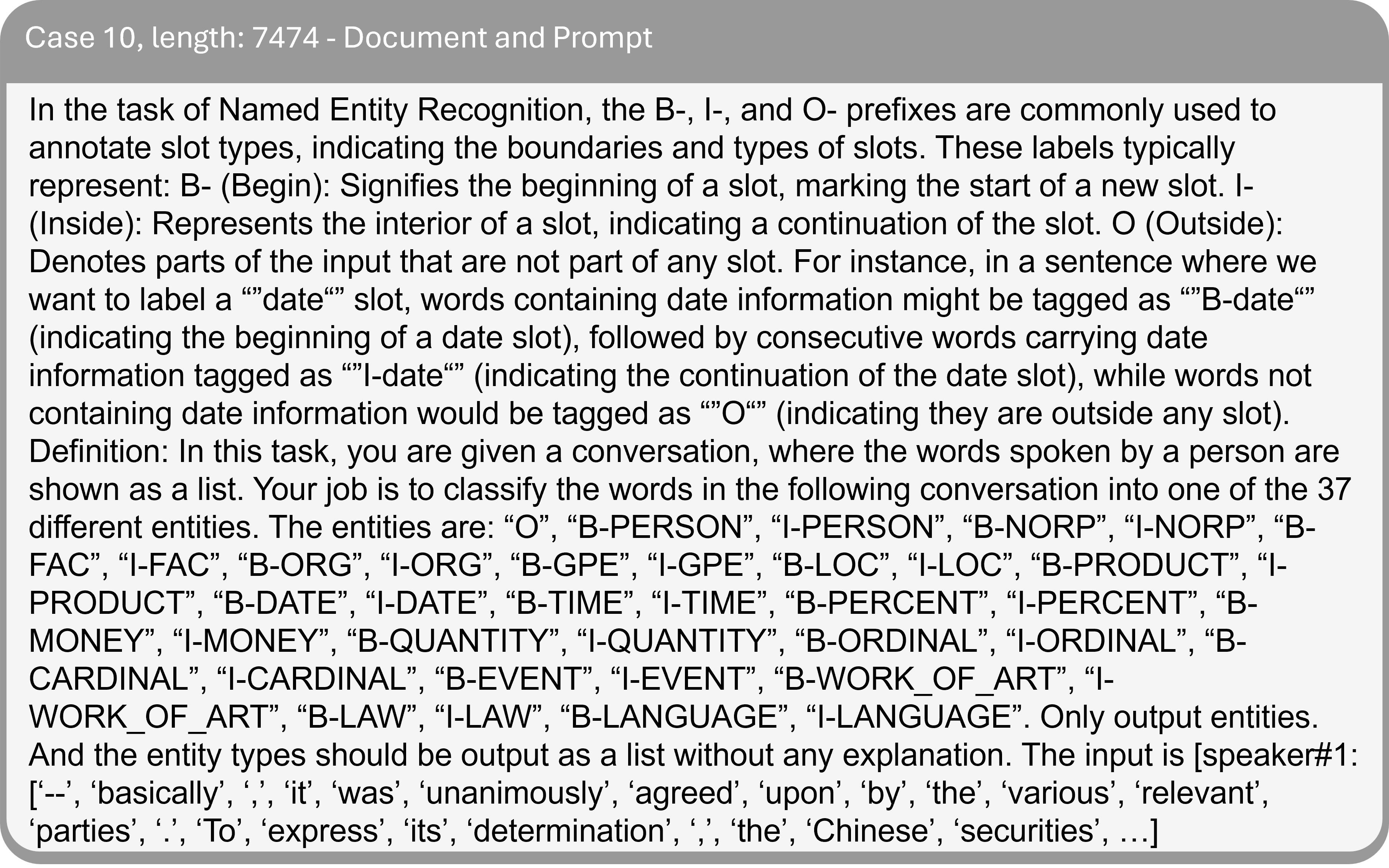} 
    \end{subfigure}
    \vspace{1pt}
    \begin{subfigure}[b]{1.0\linewidth}  
        \includegraphics[width=\textwidth]{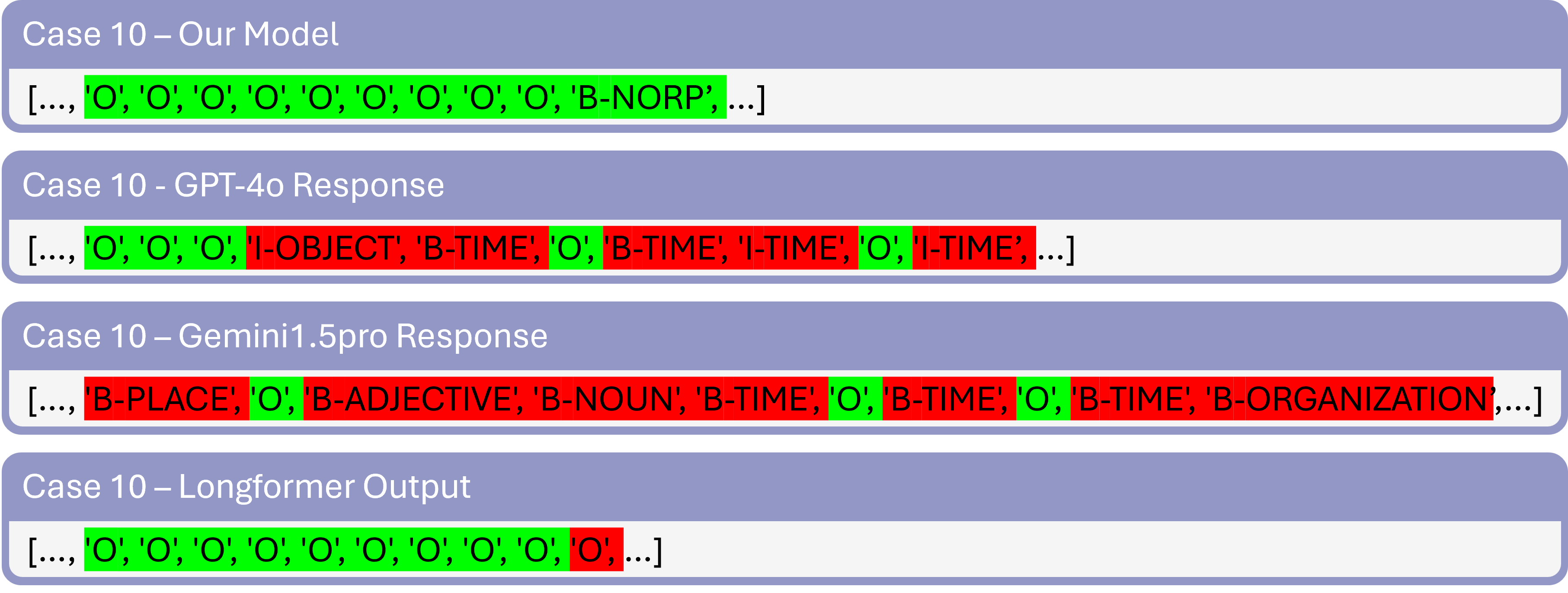}
    \end{subfigure}
    \caption{Prompt and output for a sample document of length 7474 in \textbf{CoNLL} dataset for NER task, where correct predictions are highlighted in green and wrong predictions are highlighted in red.}
    \label{case10}
\end{figure}

\begin{figure}[!h]
    \centering
    \begin{subfigure}[b]{1.0\linewidth}   
        \includegraphics[width=\textwidth]{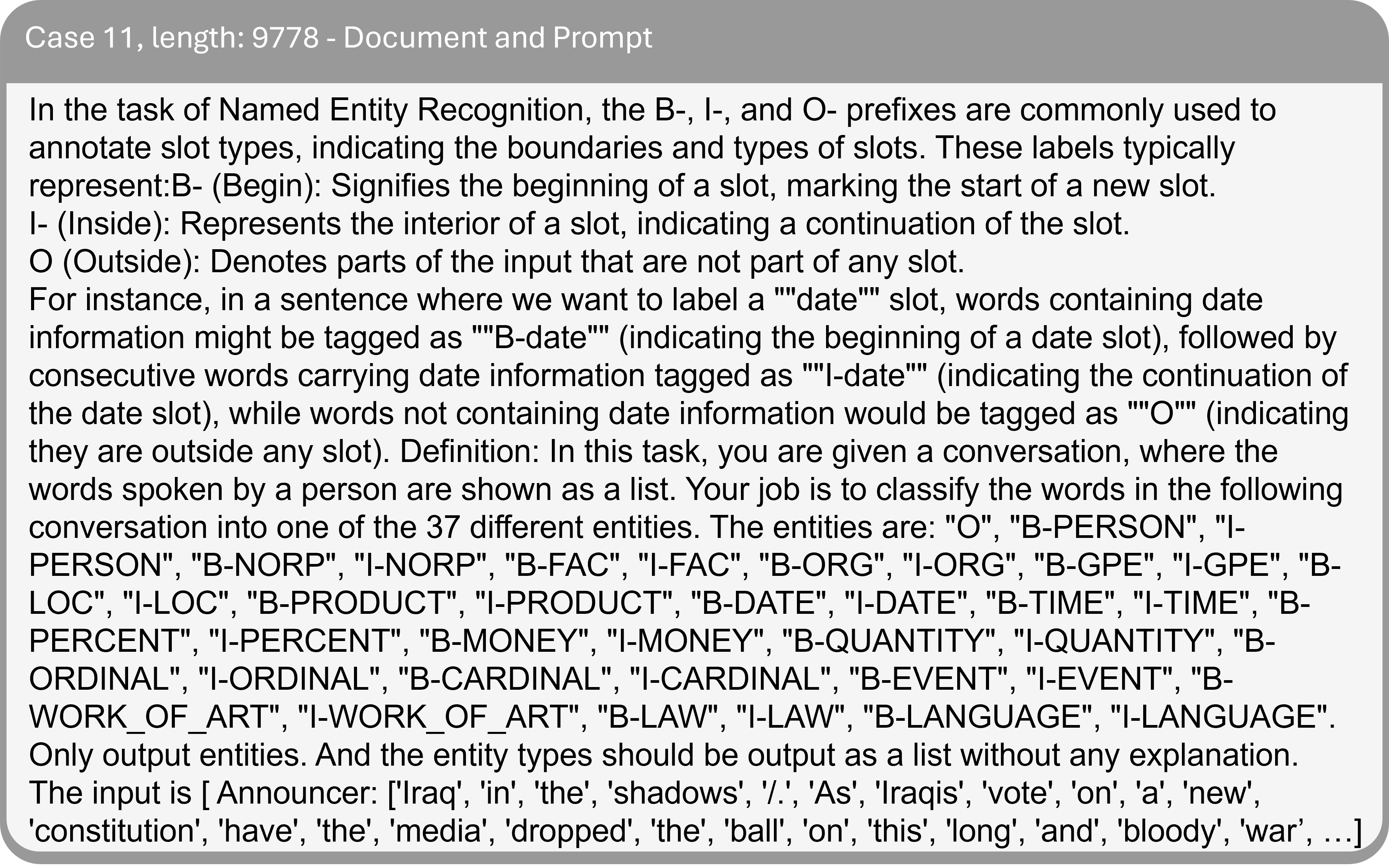} 
    \end{subfigure}
    \vspace{1pt}
    \begin{subfigure}[b]{1.0\linewidth}  
        \includegraphics[width=\textwidth]{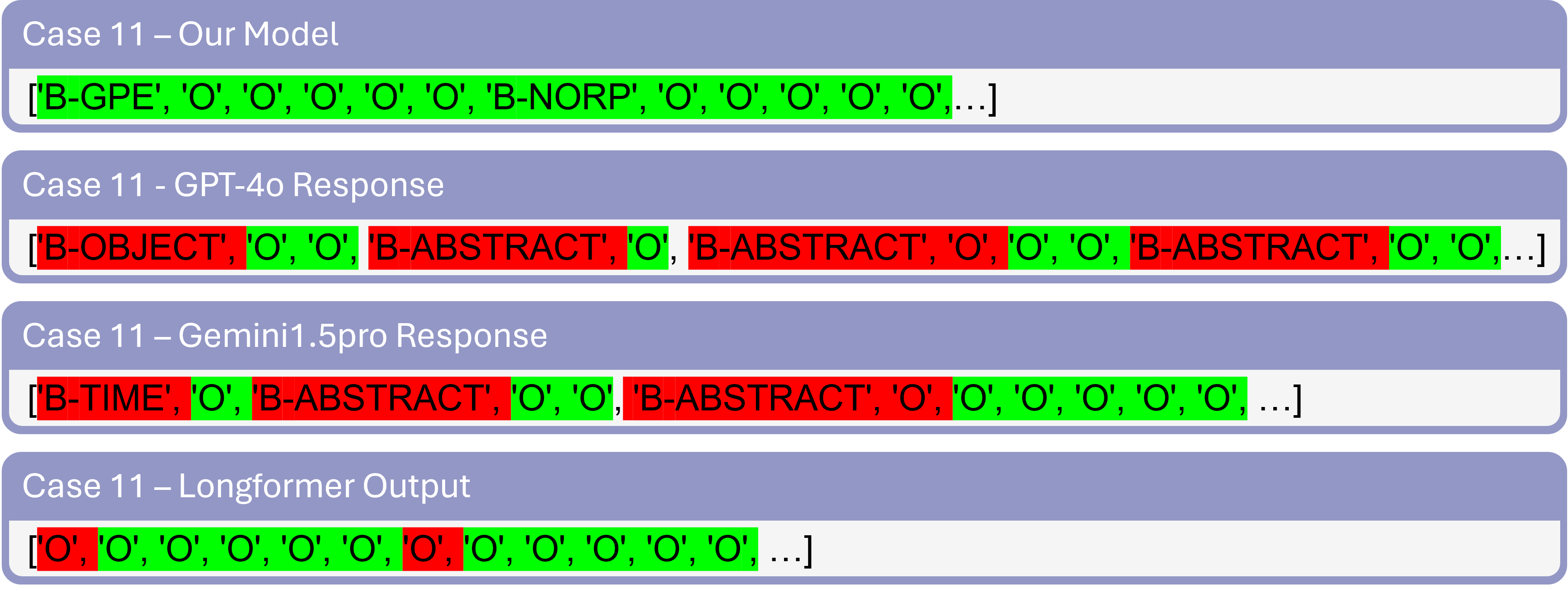}
    \end{subfigure}
    \caption{Prompt and output for a sample document of length 9778 in \textbf{CoNLL} dataset for NER task, where correct predictions are highlighted in green and wrong predictions are highlighted in red.}
    \label{case11}
\end{figure}

In this section, we will present several prompt and output samples for the long documents from the LUN (Figures~\ref{fig:case1}) and ~\ref{fig:case2}) and Hyperpartisan (Figures~\ref{fig:case3} and ~\ref{fig:case4}) datasets for document classification, as well as GUM (Figures~\ref{case5}, \ref{case6} and ~\ref{case7}) and CoNLL (Figures~\ref{case8}, ~\ref{case9}, ~\ref{case10} and ~\ref{case11}) datasets for NER task. Documents with various lengths are randomly selected to see the comparison of our model against GPT-4, Gemini1.5pro and Longformer. While there is always at least one baseline which predicts wrongly for the difficult cases presented for the document classification task, we can observe that our model consistently classifies those documents well. For the token classification task, our model can also correctly classify more tokens than each baseline across the shown cases.

\end{document}